\documentclass[10pt,journal,compsoc]{IEEEtran}

\usepackage{times}
\usepackage{epsfig}
\usepackage{graphicx}
\usepackage{amsmath}
\usepackage{amssymb}

\usepackage{comment}
\usepackage{hhline}

\usepackage{caption} 
\usepackage[export]{adjustbox}
\usepackage{tabularx}

\usepackage{enumitem}

\usepackage{textcomp}

\usepackage[pagebackref=true,breaklinks=true,colorlinks,bookmarks=false]{hyperref}

\ifCLASSOPTIONcompsoc
  \usepackage[nocompress]{cite}
\else
  \usepackage{cite}

\fi

\ifCLASSINFOpdf
 
\else
 
\fi

\begin{document}

\title{Distribution Matching for Heterogeneous Multi-Task Learning: a Large-scale Face Study}
%\title{Knowledge Distillation for Heterogeneous Multi-Task Learning: a Large-scale Face Study}
%Face Behavior \`a la carte: Expressions, Affect and Action Units in a Single Network}

\author{Dimitrios~Kollias, %~\IEEEmembership{Member,~IEEE,}
        Viktoriia~Sharmanska %,~\IEEEmembership{Fellow,~OSA,}
         and~Stefanos~Zafeiriou%~\IEEEmembership{Life~Fellow,~IEEE}% <-this % stops a space
\IEEEcompsocitemizethanks{\IEEEcompsocthanksitem D. Kollias is with the School of Computing and Mathematical Sciences, University of Greenwich, London, UK.\protect\\
E-mail: D.Kollias@greenwich.ac.uk
\IEEEcompsocthanksitem V. Sharmanska is with the Department of Informatics, University of Sussex, UK and with the Department of Computing, Imperial College London, UK.% <-this % stops an unwanted space
\IEEEcompsocthanksitem S. Zafeiriou is with the Department of Computing, Imperial College London, UK.
}}

\markboth{IEEE Transactions on Pattern Analysis and Machine Intelligence}%
{}

\IEEEtitleabstractindextext{%
\begin{abstract}

 Multi-Task Learning (MTL) has emerged as a methodology in which multiple tasks are jointly learned by a shared learning algorithm, such as a deep neural network. MTL is based on the assumption that the tasks under consideration are related; therefore it exploits shared knowledge for improving performance on each individual task. Tasks are generally considered to be homogeneous, i.e., to refer to the same type of problem, e.g., classification. Moreover, MTL is usually based on ground truth annotations with full, or partial overlap across tasks; i.e., for each input sample, there exist annotations for all or most of the tasks. 
In this work, we deal with heterogeneous MTL, simultaneously addressing detection, classification and regression problems. We explore task-relatedness as a means for co-training, in a weakly-supervised way, tasks that contain little, or even non-overlapping annotations. Task-relatedness is introduced in MTL, either explicitly through prior expert knowledge, or through data-driven studies. We propose a novel distribution matching approach, in which knowledge exchange is enabled between tasks, via matching of their predictions' distributions. Based  on  this approach,  we  build FaceBehaviorNet, the first framework for  large-scale face analysis, by jointly learning all facial behavior tasks. We develop case studies for: i) continuous affect estimation, facial action unit detection and basic emotion recognition;
ii) facial attribute detection and face identification. 
 We illustrate that co-training via task relatedness alleviates negative transfer, i.e., cases in which MT model's performance is, in some task(s), worse than that of a single-task model. Since FaceBehaviorNet learns features that encapsulate all aspects of facial behavior, we conduct zero- and few-shot learning to perform tasks beyond the ones that it has been trained for, such as compound emotion recognition. 
By conducting a very large experimental study, utilizing 10 databases, we illustrate that our approach outperforms, by large margins, the state-of-the-art in all tasks and in all databases, even in these which have not been used in its training.

\smallskip

\end{abstract}

}

\maketitle

\IEEEdisplaynontitleabstractindextext

\IEEEpeerreviewmaketitle

\IEEEraisesectionheading{\section{Introduction}\label{sec:introduction}}

\IEEEPARstart{H}{olistic} frameworks, where several learning tasks are interconnected and explicable by the reference to the whole, are common in computer vision. A diverse set of examples includes a scene understanding framework that reasons about 3D object detection, semantic segmentation and depth reconstruction \cite{wang2015holistic}, a face analysis framework that addresses face detection, landmark localization, gender recognition, age estimation \cite{ranjan2017all}, a universal network for low-, mid-, high-level vision \cite{kokkinos2017ubernet}, a large-scale framework of visual tasks for indoor scenes \cite{zamir2018taskonomy}. 
Most if not all prior works rely on building a multi-task framework where learning is done based on the ground truth annotations with full or partial overlap across tasks. During training, all the tasks are optimised simultaneously aiming for representation learning that supports a holistic view of the framework. 
Our approach for building such framework falls under heterogeneous multi-task learning \cite{zhang2021survey,pan2010survey}. As opposed to standard multi-task learning with a single type of e.g. classification tasks, heterogeneous learning addresses jointly different types of supervised tasks including classification, detection, and regression problems, which poses new challenges for effective knowledge transfer. 

What makes our work different from the previous holistic approaches is exploring the idea of task-relatedness as means for co-training the heterogeneous tasks. 
In our work, relatedness between heterogeneous tasks is either provided explicitly in a form of expert knowledge, or is inferred based on empirical studies. 
Importantly, in co-training, the related tasks exchange their predictions and iteratively teach each other so that predictors of all tasks can excel \emph{even if we have limited or no data} for some of them.
How to perform knowledge exchange and co-training between heterogeneous tasks that could generalize to different scenarios and across datasets is the key challenge that we address in this work. 
We propose an effective distribution matching approach based on distillation \cite{hinton2015distilling}, where knowledge exchange between tasks is enabled via distribution matching over their predictions.  
%that is provided explicitly either as expert knowledge or from empirical evidence. 
%In this form, it is similarly motivated to the classical multi-task literature exploring feature sharing \cite{argyriou2007multi} and task relatedness \cite{jayaraman2014decorrelating} during training; more examples can be found in the surveys \cite{zhang2017survey,pan2010survey}. However in the multi-task setting, one typically assumes homogeneity of the tasks, i.e. tasks of the same type such as object classifiers or attribute detectors. 
%
Based on this approach, we build the first holistic framework for large-scale face analysis, FaceBehaviorNet, with case studies in affective computing and in face recognition. 
In affective computing, we have heterogeneous tasks such as predicting categorical emotions (e.g. happy, sad, angry, surprised), predicting continuous dimensions of valence and arousal (how positive/negative, active/passive the emotional state is), predicting activations of binary action units \cite{ekman1997face} (activation of facial muscles) to explain the human's affective state. 
In face recognition, we consider the two interconnected tasks of face identification and facial attribute detection in a single holistic framework. 
%The main difference and novelty of our work is that the proposed holistic framework (i) explores the relatedness of heterogeneous tasks, e.g. classification, \textcolor{red}{attribute} detection, regression; (ii) operates over datasets with partial or non-overlapping annotations of the tasks; (iii) encodes explicit relationship between tasks to improve transparency and to enable expert input; \textcolor{red}{(iv) negates the effect of negative transfer that occurs in multi-task learning}.  

Up until now training holistic models has been primarily addressed by combining multiple datasets to solve individual tasks \cite{ranjan2017all}, or by collecting the annotations in terms of all tasks \cite{zamir2018taskonomy, kokkinos2017ubernet}. 
As an example, let us consider facial behavior analysis which is one of our case studies. A lot of effort has been made towards collecting datasets of naturalistic facial behavior captured in uncontrolled conditions, \emph{in-the-wild} \cite{kollias2018deep, zafeiriou2017aff, mollahosseini2017affectnet, emotionet2016}. 
Among the three heterogeneous tasks in affective computing, collecting annotations of action units is particularly costly, as it requires skilled annotators to perform the task. Nevertheless there has been a lot of effort to collect those annotations and develop automatic toolboxes \cite{emotionet2016, openface2015}. 
%There is a rich literature on recognition of basic emotion categories or expressions \cite{ekman1971constants} such as anger, disgust, fear, happiness, sadness, surprise and neutral in-the-wild \cite{dalgleish2000handbook,cowie2003describing}. Recently, continuous affect dimensions such as valence (how positive/negative a person is) and arousal (how active/passive a person is) (VA) have attracted attention, as they are naturally suited to represent emotional state and how it changes over time. %Datasets for continuous affect are simpler to collect while benefiting from human computer interactions techniques. 
%Automatic facial analysis has been also studied in terms of the facial action units (AUs) coding system  \cite{ekman1997face}. This system is a systematic way to code the facial motion with respect to activation of facial muscles. %It has been widely adopted as a common standard towards systematically categorising physical manifestation of complex facial expressions. 
%The dataset collection of action units is very costly, as it requires skilled annotators to perform the task. Nevertheless there has been a lot of effort to collect action unit annotations and develop automatic AUs annotation toolboxes \cite{emotionet2016, openface2015}. 
%
The datasets collected so far have annotations for training some of the heterogeneous tasks and despite significant effort \cite{kollias2019expression}, there is no dataset that for each image or video has complete annotations of all three tasks. 
Co-training via task relatedness is an effective way of aggregating knowledge across datasets and transferring it across heterogeneous tasks, especially with little or non-overlapping annotations. 
%
%In contrast to previous efforts of building general purpose holistic frameworks, where the tasks have an associated task-specific loss, we propose to explore task-relatedness as a supervisory signal for co-training the heterogeneous tasks.

In this work we discuss two strategies to infer task relatedness: i) via domain knowledge, ii) via dataset annotation, see Table~\ref{table:EmoAUs}, Table~\ref{relate_attr_id} in our case studies.  
For example, the three aforementioned tasks of facial behavior analysis are interconnected with known strengths of relatedness from the literature. In \cite{ekman1997face}, the facial action coding system (FACS) has been built to indicate for each of the basic expressions its \emph{prototypical} action units. In \cite{du2014compound}, a dedicated user study has been conducted to study the relationship between AUs activations and emotion expressions for basic types and beyond -- see Table~\ref{table:EmoAUs}. 
In \cite{khorrami2015deep}, the authors show that neural networks trained for expression recognition implicitly learn action units. 
Also, in \cite{mehu2015emotion} the authors have discovered that valence and arousal dimensions could be interpreted by AUs. For example, AU12 (lip corner puller) is related to positive valence.  
In our second case study on face recognition, we have an example of a dataset such as CelebA \cite{liu2015faceattributes}, where annotations for both tasks, identification and attribute prediction, are available for each image. %The latter one was collected using professional annotation agency. 
We can infer task relatedness based on the annotations empirically -- see examples in Table~\ref{relate_attr_id}. 

One of the important challenges in multi-task learning is how to avoid negative transfer, when the performance of the multi-task model can be worse than that of a single-task model \cite{wang2019characterizing,liu2019loss}. Negative transfer occurs naturally in the multi-task learning scenarios when: i) source data are heterogeneous or less related (as all tasks are diverse to each other, there is no suitable common latent representation and thus multi-task learning produces poor representations); ii) one task dominates the training process (in this scenario, one group of related tasks may dominate the training process and negative transfer may occur simultaneously on tasks outside the dominant group). 
We demonstrate empirically that the proposed distribution matching approach based on task relatedness can alleviate the problem of negative transfer in FaceBehaviorNet. 

Our main contributions are as follows: 
\begin{itemize}[leftmargin=*,noitemsep,nolistsep]
\item We propose a flexible holistic framework that can accommodate heterogeneous tasks with encoding prior knowledge of tasks relatedness. In our experiments we evaluate two effective strategies of task relatedness: 
a) obtained from domain knowledge, e.g. based on a cognitive study \cite{du2014compound}, and b) inferred empirically from dataset annotations. 
\item We propose an effective weakly-supervised learning approach that couples, via distribution matching and label co-annotation, heterogeneous tasks which contain little, or even non-overlapping annotations; we show its effectiveness for face analysis in two case studies: affective computing and face recognition. 
\item We present the first, to the best of our knowledge, holistic network for facial behavior analysis; this is capable of simultaneously predicting 7 basic expressions, 17 action units and continuous valence-arousal emotion dimensions in-the-wild. For network training and evaluation we utilize publicly available in-the-wild databases.  %For network training we utilize all publicly available in-the-wild databases that, in total, consist of over 5M images with partial and/or non-overlapping annotations for different tasks. 
This network will be made publicly available. All available databases are automatically annotated for all tasks by this network; these annotations will also be made publicly available.
\item We conduct an extensive experimental study, currently the largest to the best of our knowledge, in which we evaluate FaceBehaviorNet on 10 databases and compare its performance to single-task networks. We demonstrate that FaceBehaviorNet when trained with the proposed coupling losses \textit{greatly outperforms single-task networks in all tasks and in all databases}, even in ones that have not been used in its training. This validates that the network’s capabilities are  enhanced  when  it  is  jointly  trained  for  all  related  tasks. It is also shown that the distribution matching approach for knowledge distillation across heterogeneous tasks successfully prevents negative transfer in multi-task learning.
We also show that FaceBehaviorNet displays the best performance \textit{across all 10 databases} outperforming the-state-of-the-art methods in each database.

%\textcolor{red}{We show that FaceBehaviorNet greatly outperforms each of the single-task networks, validating that our network's capabilities are enhanced when it is jointly trained for all related tasks. Furthermore, the distribution matching approach for knowledge distillation across heterogeneous tasks successfully prevents negative transfer in multi-task learning.}

\item We further explore the feature representation learned in the joint training and show its generalization abilities on the task of compound expression recognition when no or little training data is available (zero-shot and few-shot learning). 
\end{itemize}
% extensive experiments using 2M images, 7 databases; testing 2 new databases; across all images; scale of the study - nobody has done this; cross dataset
% general approach two case studies that show different coupling in multi-task setting
% zero shot into methodology
% distillation
% 
\section{Related work}\label{related_work}

Works exist in literature that use emotion labels to complement missing AU annotations or increase generalization of AU classifiers \cite{ruiz2015emotions,yang2016multiple,wang2017expression}. Our work deviates from such methods, as we target joint learning of all three facial behavior tasks via a single holistic framework, whilst these works perform  only AU detection (and not emotion recognition and valence-arousal estimation).

Multi-task learning (MTL) was first studied in \cite{caruana1997multitask}, where the authors proposed to jointly learn parallel tasks sharing a common representation; they used part of the knowledge learned when solving one task, so as to improve learning of the other related tasks. Since then, several approaches have adopted MTL for solving different problems in computer vision and machine learning. In the face analysis domain, the use of MTL is somewhat limited. In \cite{jang2019registration}, Face-SSD was proposed for
detecting faces and performing various -separate and independent- face-related tasks, including
smile recognition, face attribute prediction and valence-arousal estimation; of these tasks just one is implemented at a time and there are no MTL experiments. The authors just mention that MTL can be used in Face-SSD and the network's optimization  loss  function should be  the  sum  of  the independent task losses.

In \cite{wang2017multi}, MTL was tackled through a neural network that jointly handled face recognition and facial attribute prediction tasks. At first the authors trained a network for facial attribute detection on CelebA; then they used it to generate attribute labels for another database (CASIA-WebFace) that only contained facial identification (id) labels. Finally, the authors trained a multi-task network for attribute detection and face identification on that database (that contained the id labels and the generated attribute labels). The loss function of the network was the sum of the independent task losses.
%MTL helped capture global feature and local attribute information simultaneously.

An approach targeting a problem similar to ours is \cite{cui2020knowledge}, presenting  a knowledge augmented deep learning framework for joint AU detection and facial expression recognition. The described framework consisted of a knowledge model - represented by a Bayesian Network - and three neural network based sub-models. An image-based FER model  performed facial expression classification directly from image data. An AU model performed AU detection from the images. The knowledge model was used to weakly supervise the learning of the AU detector. An AU-based FER model performed expression recognition from AU detection results; it was mainly introduced  to assist the  model integration process. The three neural network models were initially trained independently and they were then refined jointly until convergence. 
It should be mentioned that this work does not cover valence-arousal estimation. Moreover, it utilizes only highly controlled databases for training and evaluation; there is no proof that it can be applied effectively to real world in-the-wild data cases. In addition, it does not generate a single network, but three distinct ones; having high time and space requirements and high computational complexity.

Another work with a goal similar to ours is \cite{deng2020multitask}, where a unified model performing facial action unit detection, expression classification, and valence-arousal estimation was proposed as part of the ABAW Competition at IEEE FG 2020\footnote{https://ibug.doc.ic.ac.uk/resources/fg-2020-competition-affective-behavior-analysis/}. The authors used the Aff-Wild2 database \cite{kollias2019expression,kollias2020analysing} that contains annotations for all 3 tasks. However not all images were annotated for all tasks. To tackle the incomplete labels' problem, the authors at first trained a teacher multi-task model using only the given complete labels. Then by testing that network on the database, they generated annotations (soft labels) for the missing labels. Finally they trained a student multi-task network on the union of the original and soft labels; that network outperformed the teacher one. They also applied data balancing techniques and developed ensembles for further boosting the performance. 
The teacher model did not take into account the fact that the three tasks are interconnected - they simply used an overall loss equal to the sum of the independent task losses. Thus, the student model did not learn this relatedness. It should also be added that this work utilized and was evaluated on only one database, i.e., the Aff-Wild2; the soft labels generated by  the teacher model were not reliable.

In \cite{kuhnke2020two} a multi-task and multi-modal network was proposed for valence-arousal estimation, action unit detection and seven basic expression classification. The authors utilized a 3D ResNet for processing the image modality (video frames) and a ResNet for processing the audio modality (Mel-spectrograms). The features from these two networks were concatenated and fed to a fully connected layer that provided the final estimates for the three tasks. However, this work was trained with and was evaluated on the Aff-Wild2 database, utilizing only the overlapping and complete annotations.

\begin{table*}[t]
\caption{Relatedness between: i) basic emotions and their prototypical and observational AUs from \cite{du2014compound}: the weights $w$ in brackets correspond to the fraction of annotators that observed the AU activation; ii) basic emotions and AUs, inferred from Aff-Wild2: the weights $w$ in brackets correspond to the percentage of images annotated with the specific expression in which the AU was activated.}
\label{table:EmoAUs}
\centering
\scalebox{1.}{
\begin{tabular}{|l|c|c|c|}
\hline
 & \multicolumn{2}{c|}{\begin{tabular}{@{}c@{}}    Cognitive-Psychological  Study \cite{du2014compound} \end{tabular}} & Empirical  Evidences, Aff-Wild2 \\
\hline
Emotion   & Prototypical AUs & Observational AUs (with weights $w$) & AUs (with weights $w$)\\
\hline\hline
happiness &  12, 25 & 6 (0.51)  & 12 (0.82), 25 (0.7), 6 (0.57), 7 (0.83), 10 (0.63) \\
\hline
sadness &  4, 15 & 1 (0.6), 6 (0.5), 11 (0.26), 17 (0.67) & 4 (0.53), 15 (0.42), 1 (0.31), 7 (0.13), 17 (0.1) \\
\hline
fear &  1, 4, 20, 25 &2 (0.57), 5 (0.63), 26 (0.33) & 1 (0.52), 4 (0.4), 25 (0.85), 5 (0.38), 7 (0.57), 10 (0.57) \\
\hline
anger &4, 7, 24 &10 (0.26), 17 (0.52), 23 (0.29) & 4 (0.65), 7 (0.45), 25 (0.4), 10 (0.33), 9 (0.15)\\
\hline
surprise &1, 2, 25, 26 &5 (0.66)  & 1 (0.38), 2 (0.37), 25 (0.85), 26 (0.3), 5 (0.5), 7 (0.2) \\
\hline
disgust &9, 10, 17 & 4 (0.31), 24 (0.26) & 9 (0.21), 10 (0.85), 17 (0.23), 4 (0.6), 7 (0.75), 25 (0.8)\\
\hline
\end{tabular}
}
\end{table*}

In \cite{wang2018two}, a two-level attention with two stage multi-task learning framework was constructed for emotion recognition and valence-arousal estimation. In the first attention level, a CNN extracted position-level features and then in the second an RNN with self-attention was proposed to model the relationship between layer-level features. This work utilized the AffectNet database \cite{mollahosseini2017affectnet}, annotated for both tasks and thus only containing overlapping and complete annotations. In addition, this work did not tackle the action unit detection task.

\section{The Proposed Approach}\label{approach}

Let us consider a set of \textit{m} tasks $ {\{ \mathcal{T}_i \}}^m_{i=1}$. 
In task $\mathcal{T}_i$, the observations are generated by the underlying distribution $\mathcal{D}_i$ over inputs $\mathcal{X}$ and their labels $\mathcal{Y}$ associated with the task. %the real target is determined by the underlying labelling
%functions $\ell_i: \mathcal{X} \rightarrow \mathcal{Y}$ for $ {\{ ( \mathcal{D}_i , \ell_i) \}}^m_{i=1}$. %Let us also assume that each task \textit{t} has $m_t$ examples, with $\sum_{t=1}^T{m_t} = m$.
For the \textit{i}-th task $T_i$, the training set ${D}_i$ consists of $n_i$ data points $ (\mathbf{x}_j^i, y_j^i),$ $j= 1,\ldots,n_i$ with $\mathbf{x}_j^i \in \mathbb{R}^d$ and its corresponding output $y_j^i \in \mathbb{R}$ if it is a regression task, or $y_j^i \in \{0,1\}$ if it is a binary classification task, or $y_j^i \in \{0,1\}^k$ (one-hot encoding) if it is a (mutually exclusive) k-class classification task.

Then, the goal of MTL is to find \textit{m} hypothesis: $h_1, . . . , h_m$ over the hypothesis space $\mathcal{H}$ to control the average expected error over all tasks: $ \frac{1}{m} \sum_{i=1}^{m}{\mathbb{E}_{(\mathbf{x},y)  \sim \mathcal{D}_i}} \mathcal{L}(h_i(\mathbf{x}), y)$ with $\mathcal{L}$ being the loss function. We can also define a weight $\mathbf{w_i} \in \Delta^m$, ${\{ \mathbf{{w}}_{i} \}}^m_{i=1} > 0$ to govern the contribution of each task. Then the overall loss is:
\begin{align}
 \mathcal{L}_{MT} &= \frac{1}{m} \sum_{i=1}^{m}{ \mathbf{w_i} \cdot \mathbb{E}_{(\mathbf{x},y)  \sim \mathcal{D}_i}} \mathcal{L}(h_i(\mathbf{x}),y). %\ell_i(\mathbf{x})) 
 %\nonumber \\
% &= \frac{1}{m} \sum_{i=1}^{m}{ \mathbf{w_i} \cdot      \frac{1}{m_i} \sum_{j=1}^{m_i}{    \mathcal{L}(h_i(\mathbf{x}_j^i), \ell_i(\mathbf{x}_j^i))  }  } \nonumber \\
% &= \frac{1}{m} \sum_{i=1}^{m}{ \mathbf{w_i} \cdot      \frac{1}{m_i} \sum_{j=1}^{m_i}{    \mathcal{L}(h_i(\mathbf{x}_j^i), y_j^i )  }  } 
\label{eq:ccc}
\end{align}

If it is a regression task, the loss $ \mathcal{L}$ can take a form of MAE, MSE or a correlation-based loss. If it is a binary classification task, the loss $ \mathcal{L}$ can be binary/sigmoid cross entropy loss. If it is a (mutually exclusive) k-class classification task, the loss $ \mathcal{L}$ can be softmax cross entropy loss. In the case of a neural network the hypothesis can be expressed as $f(\{\boldsymbol{\theta}\}, \mathbf{x})$ where $\{\boldsymbol{\theta}\}$ denotes the set of weights of the neural network learned during training.

%In the case of a neural network each hypothesis $h_i$ can be expressed as $f_i(\{\boldsymbol{\theta}\}, \mathbf{x})$ where $\{\boldsymbol{\theta}\}$ denotes the set of weights of the neural network learned during training.

In the following, we present the proposed framework via two examined case studies. The framework includes inferring the relationship between the tasks (either via domain knowledge or dataset annotation) and using it for coupling them during MTL. The coupling is achieved via the proposed co-annotation and distribution matching losses. These losses %are architecture agnostic; they 
can be incorporated and used in any deep neural network that performs MTL.

\subsection{Case Study I: Affective Computing}

We start with the multi-task formulation of the facial behavior model. 
In this model we have three objectives: (1) learning seven basic emotions, 
(2) detecting activations of $17$ binary facial action units, (3) learning the intensity of the valence and arousal continuous affect dimensions. 
We train a multi-task neural network model to jointly perform (1)-(3). 
For a given image $x \in \mathcal{X}$, we can have label annotations of either one of seven basic emotions $y_{emo} \in \{1,2,\ldots,7\}$, or $17$\footnote{In fact, $17$ is an aggregate of action units in all datasets; typically each dataset has from 10 to 12 AUs labelled by purposely trained annotators.} binary action units activations $y_{au} \in \{0,1\}^{17}$, or two continuous affect dimensions, valence and arousal, $y_{va} \in [-1,1]^{2}$.
For simplicity of presentation, we use the same notation $x$ for all images leaving the context to be explained by the label notations.  
We train the multi-task model by minimizing the following objective:
%\vspace*{-4px}
\begin{align}
\mathcal{L}_{MT} &= \lambda_{1} \mathcal{L}_{Emo} + \lambda_{2} \mathcal{L}_{AU} + \lambda_{3} \mathcal{L}_{VA} \label{eq:mt1}\\
\mathcal{L}_{Emo} &= \mathbb{E}_{x,y_{emo}}\big[-\text{log } p (y_{emo}|x)\big]\nonumber\\ 
\mathcal{L}_{AU} &= \mathbb{E}_{x,y_{au}}\big[- \text{log } p (y_{au}|x)\big]\nonumber\\ 
\mathcal{L}_{VA} &= 1- CCC(y_{va},\bar{y}_{va}),\nonumber
\end{align}
where the first term is the cross entropy loss computed over images with a basic emotion label, the second term is the binary cross entropy loss computed over images with $17$ AUs activations, 

\noindent $\text{log } p (y_{au}|x) :=  \frac{\sum_{i=1}^{17} \delta_i \cdot [y_{au}^i\text{log } p (y_{au}^i|x) + (1-y_{au}^i)\text{log } (1-p (y_{au}^i|x))]} {\sum_{k=1}^{17} \delta_k} $, 

\smallskip
\noindent where $\delta_i \in \{0,1\}$ indicates whether the image contains annotation for $AU_i$.
The third term measures the concordance correlation coefficient between the ground truth valence and arousal $y_{va}$ and the predicted $\bar{y}_{va}$, $CCC(y_{va},\bar{y}_{va})=\frac{\rho_a + \rho_v}{2}$, where for $i \in \{v,a\}$, $y_i$ is the ground truth, $\bar{y_i}$ is the predicted value and  $\rho_i=$
\begin{align}
 \frac{2\cdot\mathbb{E}\big[(y_{i}-\mathbb{E}_{y_{i}})\cdot(\bar{y}_{i}-\mathbb{E}_{\bar{y}_{i}})\big]}
{\mathbb{E}^2\big[{(y_{i}-\mathbb{E}_{y_{i}}})^2\big] + \mathbb{E}^2\big[{(\bar{y}_{i}-\mathbb{E}_{\bar{y}_{i}}})^2\big] + (\mathbb{E}_{y_{i}} -\mathbb{E}_{\bar{y}_{i}})^2}.\nonumber
\label{eq:ccc}
\end{align}

\subsubsection{\textbf{Task-Relatedness}}
In the seminal work of \cite{du2014compound}, the authors conduct a study on the relationship between emotions (basic and compound) and facial action unit activations. The summary of the study is a Table of the emotions and their prototypical and observational action units, which we include in Table \ref{table:EmoAUs} for completeness. Prototypical action units are ones that are labelled as activated across all annotators' responses, observational are action units that are labelled as activated by a fraction of annotators. For example, in emotion \emph{happiness} the prototypical are AU12 and AU25, the observational is AU6 with weight $0.51$ (observed by 51\% of the annotators). Table \ref{table:EmoAUs} provides the relatedness between emotion categories and action units obtained from this cognitive and psychological study with human participants. 

Alternatively we can infer task relatedness from external dataset annotations. In particular, we use the Aff-Wild2 database, which is the first in-the-wild database that contains annotations for all three behavior tasks to infer task relatedness. The dataset is fully annotated with basic emotions and continual emotions of valence and arousal, and a subset of it is annotated with action units. 
We first train a network for AU detection on the union of Aff-Wild2 and GFT databases \cite{girard2017sayette}, and use this network to automatically annotate Aff-Wild2 with AUs. 
Table \ref{table:EmoAUs} shows the distribution of AUs for each basic expression that we use as task relatedness for distribution matching. In parenthesis next to each AU is the percentage of images annotated with the specific expression in which this AU was activated. 
%Other means of describing task relatedness in a holistic framework will be further explored in the future.

\noindent In the following, we use the domain knowledge, specifically the cognitive and psychological study \cite{du2014compound}, to encode task relatedness and introduce the proposed approach for coupling the tasks.

%\vspace{-0.4cm}
\subsubsection{\textbf{Coupling of basic emotions and AUs}} 
\textbf{Via Co-annotation. } We propose a simple strategy of \emph{co-annotation} to couple the training of emotions and action unit predictions. 
Given an image $x$ with the ground truth %annotation of 
basic emotion $y_{emo}$, we enforce the prototypical and observational AUs of this emotion to be activated. We co-annotate the image $(x,y_{emo})$ with $y_{au}$; %that contains only the prototypical and observational AUs
 this image contributes to both 
%and include this image twice, when computing
$\mathcal{L}_{Emo}$ and $\mathcal{L}_{AU}$\footnote{Here we overload slightly our notations; for co-annotated images, $y_{au}$ has variable length and only contains prototypical and observational AUs.} in eq.  \ref{eq:mt1}. We re-weight the contributions of the observational AUs with the annotators' agreement score (from Table \ref{table:EmoAUs}).

Similarly, for an image $x$ with the ground truth  %annotation of the 
action units $y_{au}$, we check whether we can co-annotate it with an emotion label. 
{For an emotion to be present, all its prototypical and observational AUs have to be present. In cases when more than one emotion is possible, we assign the label $y_{emo}$ of the emotion with the largest requirement of prototypical and observational AUs}.
The image $(x,y_{au})$ that is co-annotated with the emotion label $y_{emo}$ %is included twice in \ref{eq:mt1}, when computing
contributes to both $\mathcal{L}_{AU}$ and $\mathcal{L}_{Emo}$ in eq. \ref{eq:mt1}. 
We call this approach the FaceBehaviorNet with co-annotation. \\
%\begin{table}[t]
%\caption{Basic emotions and their prototypical and observational AUs from \cite{du2014compound}. The weights $w$ in brackets correspond to the fraction of annotators that observed the AU activation.}
%\label{table:EmoAUs}
%\centering
%\scalebox{0.95}{
%\begin{tabular}{|l|c|c|}
%\hline
%Emotion   & Protot. AUs & Observ. AUs (with weights $w$)\\
%\hline\hline
%happiness &  12, 25 & 6 (0.51) \\
%\hline
%sadness &  4, 15 & 1 (0.6), 6 (0.5), 11 (0.26), 17 (0.67) \\
%\hline
%fear &  1, 4, 20, 25 &2 (0.57), 5 (0.63), 26 (0.33) \\
%\hline
%anger &4, 7, 24 &10 (0.26), 17 (0.52), 23 (0.29)\\
%\hline
%surprise &1, 2, 25, 26 &5 (0.66)\\
%\hline
%disguste &9, 10, 17 & 4 (0.31), 24 (0.26)\\
%\hline
%\end{tabular}
%}
%\end{table}

%\subsubsection{\textbf{Coupling of basic emotions and AUs via distribution matching}} 
\noindent\textbf{Via Distribution Matching. }
The aim here is to align the \emph{predictions} of the emotions and action units tasks during training.  
For each sample $x$ we have the predictions of emotions $p(y_{emo}|x)$ as the softmax scores over seven basic emotions and we have the prediction of AUs activations $p(y_{au}^i|x)$, $i=1,\ldots,17$ as the sigmoid scores over $17$ AUs. 

The distribution matching idea is simple: we match the distribution over AU predictions $p(y_{au}^i|x)$ with the distribution 
$q(y_{au}^i|x)$, where the AUs are modeled as a mixture over the basic emotion categories: %the prototypical and observational AUs: 
\begin{equation}
    q(y_{au}^i|x) = \sum_{y_{emo} \in \{1,\ldots,7\}} p(y_{emo}|x) \: p(y_{au}^i| y_{emo}), %{\sum_{y_{emo} \in \{1,\ldots,7\}} \: p(y_{au}^i| y_{emo})}.
\label{eq:distr}
\end{equation} 
where $p(y_{au}^i| y_{emo})$ is defined deterministically from Table~\ref{table:EmoAUs} %: $p(y_{au}^i| y_{emo})=1$ 
and is 1 for prototypical/observational action units, or 0 otherwise.  For example, AU2 is prototypical for emotion \emph{surprise} and observational for emotion \emph{fear} and thus $q(y_{\text{AU2}}|x) = p(y_{\text{surprise}}|x) + p(y_{\text{fear}}|x)$\footnote{We also tried a variant with reweighting for observational AUs, i.e. $p(y_{au}^i| y_{emo})=w$ }.  

This matching aims to make the network's predicted AUs consistent with the prototypical and observational AUs of the network's predicted emotions. So if, e.g., the network predicts the emotion \emph{happiness} with probability 1, i.e., $p(y_{\text{happiness}}|x)=1$, then the prototypical and observational AUs of \emph{happiness} -AUs 12, 25 and 6- need to be activated in the distribution q: $q(y_{\text{AU12}}|x) = 1$; $q(y_{\text{AU25}}|x) = 1$; $q(y_{\text{AU6}}|x) = 1$; $q(y_{au}^i|x) = 0$, $i \in \{1,..,14\}$. %Since the network predicts that someone is \emph{happy}, we have a belief that AUs 6, 12 and 25 could, with the above probabilistic confidence, be activated. The loss aligns this belief with the corresponding network’s AU predictions.

In spirit of the distillation approach \cite{hinton2015distilling}, we match the distributions $p(y_{au}^i|x)$ and $q(y_{au}^i|x)$ 
by minimizing the cross entropy with the soft targets loss term\footnote{This can be seen as minimizing the KL-divergence $KL(p||q)$ across the $17$ action units.}:
\begin{align}
\mathcal{L}_{DM} = \mathbb{E}_{x} \Bigg[ \sum_{i=1}^{17}[ -p(y_{au}^i|x)\text{log }q(y_{au}^i|x)] \Bigg] , \label{eq:coupleloss}
\end{align}
where all available training samples are used to match the predictions.
We call this approach FaceBehaviorNet with distr-matching.\\

%\subsubsection{\textbf{Mixing the  two  strategies,  co-annotation  and  distribution  matching}}
\noindent\textbf{Via Mixing the  two  strategies,  co-annotation  and  distribution  matching.} 
A mix of the two strategies, co-annotation and distribution matching, is also possible. Given an image $x$ with the ground truth annotation of the action units $y_{au}$, we can first co-annotate it with a \emph{soft label} in form of the  distribution over emotions and then match it with the predictions of emotions $p(y_{emo}|x)$.

More specifically, at first we compute, for each basic emotion, an indicator score, $I(y_{emo}|x)$ over its prototypical and observational AUs being present: 

\begin{align}
    I(y_{emo}|x) &= \frac{ \sum_{ y_{au},w_{au} \in \{1,\ldots,17\}} w_{au} \cdot y_{au}} {\sum_{ y_{au},w_{au} \in \{1,\ldots,17\}} w_{au}} , \text{ $y_{emo} \in \{1,\ldots,7\}$} \\ \nonumber \\
w_{au} &=\left\{  
\begin{array}{ll} 
      1, & \text{au is prototypical for $y_{emo}$ (from Table \ref{table:EmoAUs})} \\
      w, & \text{au is observational for $y_{emo}$ (from Table \ref{table:EmoAUs})} \\
      0, & \text{otherwise} \\
\end{array} 
\right. \nonumber
% \\ \nonumber \\
   % y_{emo} & \in \{1,\ldots,7\} \nonumber
%\label{eq:mix}
\end{align} 

\noindent For example, for emotion \emph{happiness}, the indicator score $I(happiness|x) = (y_{\text{AU12}} + y_{\text{AU25}} + 0.51 \cdot y_{\text{AU6}}) / (1+1+0.51)$, or all weights equal $1$ if without reweighting.

%Then, we take a softmax over the indicator scores to produce the probabilities over emotion categories; the \emph{soft} emotion label, $q(y_{emo}^i|x)$, is computed as following:
Then, we convert the indicator scores to probability scores over emotion categories; this \emph{soft} emotion label, $q(y_{emo}|x)$, is computed as following:

\begin{align}
    q(y_{emo}|x) &=  \frac{e^{I(y_{emo}|x)}}{\sum_{y'_{emo}} e^{I(y'_{emo}|x)}}
     , \text{ $ \{y_{emo}, y'_{emo}\}  \in \{1,\ldots,7\}$}
\end{align}

In this variant, every single image that has ground truth annotation of AUs will have a \emph{soft} emotion label assigned. Finally we match the predictions $p(y_{emo}|x)$ and the \emph{soft} emotion label $q(y_{emo}|x)$ by minimizing the cross entropy with the soft targets loss term:

\begin{align}
\mathcal{L}_{SCA} =  \mathbb{E}_{x} \Bigg[ \sum_{y_{emo}\in \{1,\ldots,7\}}[ -p(y_{emo}|x)\text{log }q(y_{emo}|x)] \Bigg] \label{eq:coupleloss2}
\end{align}
We call this approach FaceBehaviorNet with soft co-annotation.

\subsubsection{\textbf{Coupling of categorical emotions, AUs with continuous affect} }
In our work, continuous affect (valence and arousal) is implicitly coupled with the basic expressions and action units via a joint training procedure. Also one of the datasets we used has annotations for categorical and continuous emotions (AffectNet \cite{mollahosseini2017affectnet}). Studying an explicit relationship between them is a novel research direction beyond the scope of this work.

\subsubsection{\textbf{Compound Expressions: Zero- and Few-shot Learning}}
\label{zero-shot}

\noindent\textbf{Knowledge Generalization in Zero-shot Learning}
After FaceBehaviorNet was trained (either without or with any coupling loss or any combination of them), it was able to effectively capture and solve the three behavior tasks of valence-arousal estimation, action unit detection and basic expression classification. FaceBehaviorNet has learned features that encapsulate all aspects of facial behavior. Therefore, exploiting this fact, we describe how FaceBehaviorNet can generalize its knowledge in other emotion recognition contexts, in a zero-shot manner. For this, we use the predictions of FaceBehaviorNet together with the rules  from \cite{du2014compound} to  generate  compound  emotion  predictions.
We compute a candidate score,  $\mathcal{C}_{s}(y_{c-emo})$, for each class $y_{c-emo}$: 

\begin{align}
 \mathcal{C}_{s}(y_{c-emo}) &= \mathcal{I}_{au} + \mathcal{F}_{emo} + \mathcal{D}_{va} \\
 \mathcal{I}_{au} &=  [\sum_{k=1}^{17} p(y_{au}^k| y_{emo})] ^ {-1} \cdot  \sum_{k=1}^{17} p(y_{au}^k|x) \: p(y_{au}^k| y_{emo}) 
\nonumber\\
\mathcal{F}_{emo} &= p(y_{emo1}) + p(y_{emo2}) 
\nonumber\\
\mathcal{D}_{va}  &=\left\{  
\begin{array}{ll} 
      1, &  p(y_{v}|x) > 0\\
      0, & \text{otherwise} \\
\end{array} 
\right. \nonumber
\end{align}

%\smallskip

$\mathcal{I}_{au}$ expresses FaceBehaviorNet's predictions of only the prototypical (and observational) AUs that are associated with this compound class according to \cite{du2014compound}. In  this  manner,  every AU  acts  as  an  indicator  for  this  particular  emotion  class. This  terms  describes  the  confidence  (probability)  of  AUs that this compound emotion is present.

$\mathcal{F}_{emo}$  expresses FaceBehaviorNet's predictions of only the basic expression classes $emo1$ and $emo2$ that are mixed and form the compound class (e.g., if the compound class is happily surprised then $emo1$ is happy and $emo2$ is surprised).

$\mathcal{D}_{va}$ is added only to the happily surprised and happily disgusted classes and is either 0 or 1 depending on whether FaceBehaviorNet's valence prediction is negative or positive, respectively. The rationale is that, from all compound classes, only happily surprised and happily disgusted classes have positive valence. All other compound classes are expected to have negative valence as they correspond to negative emotions. 

The final prediction is the class that obtained the maximum candidate score.\\

%\subsubsection{\textbf{Robust Prior for Few-Shot Learning for Compound Expressions}}
\noindent\textbf{Robust Prior for Few-Shot Learning}
By having learned complex and emotionally rich features, FaceBehaviorNet can constitute a robust prior for compound emotion recognition, especially for datasets that are quite small in terms of size. Up to now, there exist only a couple of in-the-wild datasets annotated in terms of compound expressions and contain less than 5K images in total. %Additionally these datasets are very imbalanced, as there are a lot of classes (11 or 18) compared to their sizes. 
Therefore FaceBehaviorNet can be used as a pre-trained network and can be further fine-tuned (either with freezing some of its parts or not) to perform compound emotion classification.

\subsection{Case Study II: Face Recognition}%Attribute Detection \& Face Identification Case}

\begin{table*}[t]
\caption{Examples of task relatedness between identities and facial attributes inferred from CelebA}
\label{relate_attr_id}
\centering
\scalebox{1.}{
\begin{tabular}{|l||c|c|c|c|c|c|c|c|c|c|c|}
\hline
Identities   & \begin{tabular}{@{}c@{}}5 o' Clock \\ Shadow \end{tabular} & Arched Eyebrows & Attractive & \begin{tabular}{@{}c@{}}Bags \\ Under Eyes \end{tabular} & Bald & Bangs & Male & \begin{tabular}{@{}c@{}}Wearing \\ Lipstick \end{tabular} & \begin{tabular}{@{}c@{}}Wearing \\ Necklace \end{tabular} & \begin{tabular}{@{}c@{}}Wearing \\ Necktie \end{tabular} & Young  \\
\hline\hline
 \# 1 &  0.34483 & 0.03449 & 0.27586 & 0.4138 & 0. & 0. & 1. & 0. & 0. &  0. & 0.68966 \\
\hline
\# 2 &  0. & 0.125 & 0.875 & 0.125 & 0. & 0.25 & 0. & 0.5 & 0.125 & 0. & 1. \\
\hline
\# 1000 &  0. & 0.22727 &  1. & 0. & 0. &  0.04546 & 0. &  1.  & 0. & 0. & 1. \\
\hline
\# 5000 &   0.8 & 0.1 & 0.4 & 0.6 & 0.03333 & 0.06667  & 1.   & 0. & 0.  & 0.1  & 0.5   \\
\hline
\# 10000 &  0.  & 0.26667 & 1.  & 0. & 0.  & 0. & 0. & 0.76667 & 0.1 & 0.  & 1. \\
\hline
\# 10177 &   0. & 0.07692 & 0.69231 & 0.  & 0. & 0.07692 & 0. & 0.61539 & 0.15385 & 0.  & 1.  \\

\hline
\end{tabular}
}
\end{table*}

%Here we show that the approach that we described before can be additionally used for another application scenario, that of facial attribute detection and face identification.

We start with the multi-task formulation of the facial behavior model. 
In this model we have two objectives: (1) detecting activations of $40$ binary facial attributes, (2) learning to classify $10,177$ identities (ids). 
We train a multi-task neural network model to jointly perform (1) and (2).

For a given image $x \in \mathcal{X}$, we can have label annotations of either one of $10,177$ ids $y_{id} \in \{1,\ldots,10177\}$, or $40$ binary facial attributes $y_{attr} \in \{0,1\}^{40}$.
For simplicity of presentation, we use the same notation $x$ for all images leaving the context to be explained by the label notations.  
We train the multi-task model by minimizing the following objective:
%\vspace*{-4px}
\begin{align}
\mathcal{L}_{MT} &= \lambda_{1} \mathcal{L}_{Id} + \lambda_{2} \mathcal{L}_{Attr}  \label{eq:mt2}\\
%\mathcal{L}_{Emo} &= \mathbb{E}_{x,y_{id}}\big[-\text{log } p (y_{id}|x)\big]\nonumber\\ 
\mathcal{L}_{Id} &= \mathbb{E}_{x,y_{id}}\Big[ -\text{log } \frac{ e^{p(y'_{id}|x)}} {   \sum\nolimits_{y_{id}=1}^{10177}{e^{p(y_{id}|x)} } }   \Big]  \nonumber\\ 
\mathcal{L}_{Attr} &= \mathbb{E}_{x,y_{attr}}\Big[-  \sum_{i=1}^{40}  y_{attr}^i\text{log } p (y_{attr}^i|x)\Big] \nonumber \\
&+ \mathbb{E}_{x,y_{attr}}\Big[  - \sum_{i=1}^{40} (1-y_{attr}^i)\text{log } (1-p (y_{attr}^i|x)) 
\Big]\nonumber
\end{align}
where the first term is the cross entropy loss computed over images with an id label and $p(y'_{id}|x)$ is the prediction of the id that corresponds to the positive class; the second term is the binary cross entropy loss computed over images with $40$ attribute activations.

\subsubsection{\textbf{Task-Relatedness}}

There has been no relatedness study, like the one of Table \ref{table:EmoAUs}, between facial attributes and  ids. Therefore we inferred it empirically from  dataset annotations. In particular, we used the CelebA database, which contains overlapping annotations for facial attributes and identities. In other words, each image in the database is annotated both in terms of facial attributes and identities. We calculated the distribution of attributes for each identity; more precisely, for each identity (\textit{id}) and for each attribute (\textit{attr}), we divided the number of images in which the attribute \textit{attr} existed, by the total amount of images that the specific identity \textit{id} existed; as a result, we created the corresponding percentages. 

Table \ref{relate_attr_id} illustrates the relationship  between some identities and some of the 40 facial attributes. It can be seen that identities 1 and 5000 are male, whereas 2, 1000, 10000 and 10177 are female. All the images in CelebA where these 4 femals are displayed, show them young; some images in CelebA of the two male indentities show them young and some others show them old. These two male did not wear either lipstick, or necklace, in all of their images. The four female were wearing lipstick and necklace in some of their images, but not in others. Neither of the four female had a 5 o' clock shadow, whereas some images of the male had. All four female were highly attractive, whereas the two male were not that much. Mostly the two male had bags under their eyes in the images; one of them was bald in some images (probably when he was old) and he was wearing a necktie (no female was wearing a necktie).

\subsubsection{\textbf{Coupling of ids and attributes via distribution matching}} \label{distr_matching}
Since the database that we utilized contained complete and overlapping annotations, we only propose the distribution matching loss for coupling the facial attributes' and ids' predictions.
The aim is similar to the one defined and explained in Section \ref{distr_matching}; we want to align the \emph{predictions} of the ids and attributes tasks during training.  
For each sample $x$ we have the predictions of ids $p(y_{id}|x)$ as the softmax scores over 10,177 identities; we also have the prediction of attribute activations $p(y_{attr}^i|x)$, $i=1,\ldots,40$, as sigmoid scores, over $40$ AUs. 

At first, we model the attributes as a mixture over the identities,  creating distribution
$q(y_{attr}^i|x)$:
\begin{equation}
    q(y_{attr}^i|x) = \sum_{y_{id} \in \{1,\ldots,10177\}} p(y_{id}|x) \: p(y_{attr}^i| y_{id}), %{\sum_{y_{emo} \in \{1,\ldots,7\}} \: p(y_{au}^i| y_{emo})}.
\label{eq:distr2}
\end{equation} 
where $p(y_{attr}^i| y_{id})$ is defined in Table~\ref{relate_attr_id}. For example, for the '5 o' clock' attribute: $q(y_{attr}^1|x) = 0.34483 \cdot p(y_{id^1}|x) + 0.8 \cdot p(y_{i^{5000}}|x)$, assuming that for all other ids: $p(y_{attr}^i| y_{id}) = 0$.

Next, we match the distributions $p(y_{attr}^i|x)$ and $q(y_{attr}^i|x)$ 
by minimizing the cross entropy with the soft targets loss term:
\begin{align}
\mathcal{L}_{DM} = \mathbb{E}_{x} \Bigg[ \sum_{i=1}^{17}[ -p(y_{au}^i|x)\text{log }q(y_{au}^i|x)] \Bigg] , \label{eq:coupleloss3}
\end{align}
where all available training samples are used to match the predictions.
We call this approach FaceBehaviorNet with distr-matching.

\subsection{FaceBehaviorNet structure}
\begin{figure}[t]
\centering
\adjincludegraphics[height=8.7cm]{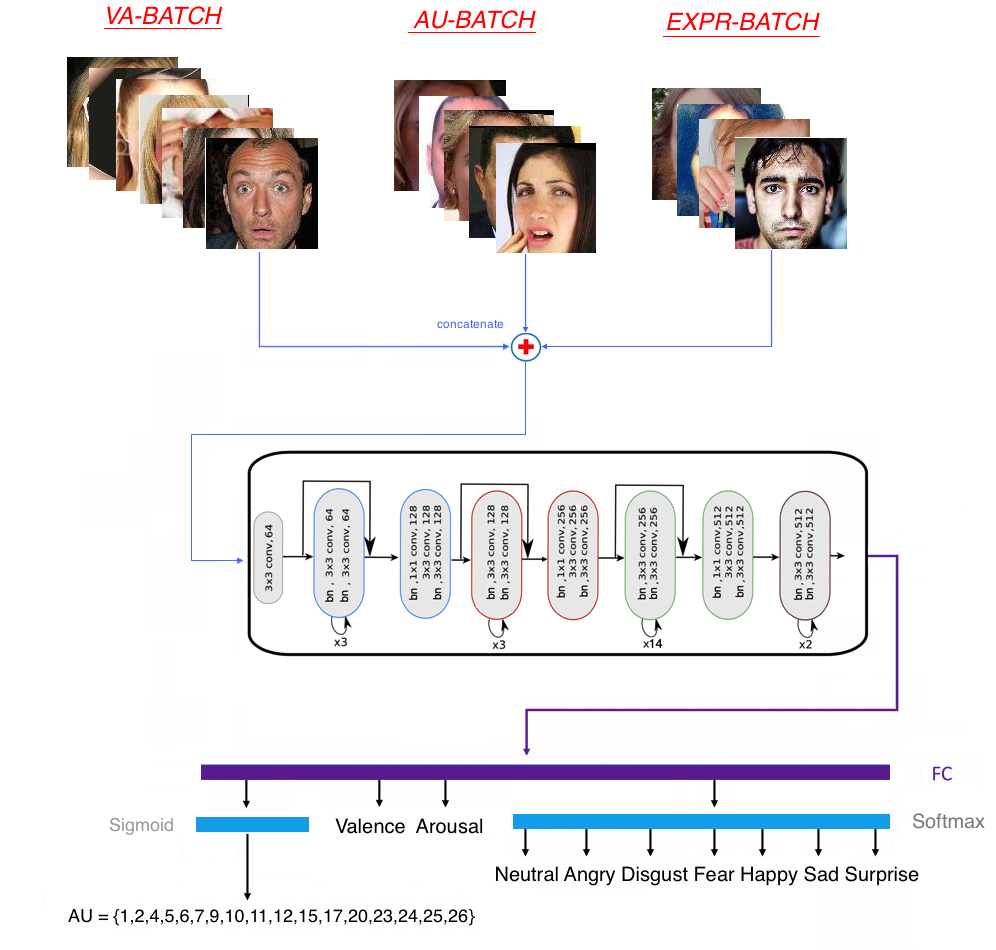} 
\caption{The holistic (multi-task, multi-domain, multi-label) FaceBehaviorNet in the affective computing case study; 'VA/AU/EXPR-BATCH' refers to batches annotated in terms of VA/AU/7 basic expressions} 
\label{facebehaviornet}
\end{figure}

Fig.\ref{facebehaviornet} shows the structure of the  holistic (multi-task, multi-domain and multi-label) FaceBehaviorNet, which is based on residual units. 'bn' stands for batch normalization, the convolution layer is in the format: filter height $\times$ filter width 'conv.', number of output feature maps; the stride is equal to 2, everywhere. In the affective computing case study: a (linear) output layer that gives final estimates for valence and arousal; it also gives 7 basic expression logits that are passed through a softmax function to get the final 7 basic expression predictions; lastly, it gives 17 AU logits that are passed through a sigmoid function to get the final 17 AU predictions. One can see that the predictions  for  all  tasks  are  pooled  from  the  same  feature  space. Fig.\ref{facebehaviornet} illustrates FaceBehaviorNet for that case. In the case study of face recognition, the (linear) output layer gives $10,177$ id logits that are passed through a softmax function to get the final id predictions and it also gives $40$ attribute logits that are passed through a sigmoid function to get the final attribute predictions.

\section{Experimental Study}

\subsection{Databases} \label{databases}

Let us first describe the databases that we utilized in all our experiments. In the affective computing case study, we selected to work with these databases because they provide a large number of samples with accurate annotations of valence-arousal, basic expressions and AUs. Training with these datasets allows our networks to learn to recognize affective states under a large number of image conditions (e.g., each database includes images at different resolutions, poses, orientations and lighting conditions). These datasets also include a variety of samples in both genders, ethnicity, races and ages.

The \textbf{Aff-Wild} database \cite{kollias2018deep} \cite{zafeiriou2017aff} has been the first large scale captured in-the-wild database, containing 298 videos (200 subjects) of around 1.25M frames, annotated in terms of valence-arousal that range in $[-1,1]$. It served as benchmark for the Aff-Wild Challenge organized in CVPR 2017. The training set consists of around 1M frames and the test set of around 216K. This database has been extented to Aff-Wild2 that consists of 545 videos with around 2.8M frames.
%%%%%%%%%%%%%%%%%%%%%%%%%%%%%%%%%%%%%%%%%%%%%%%%%%%%%%
%%%%%%%%%%%%%%%%%%%%%%%%%%%%%%%%%%%%%%%%%%%%%%%%%%%%%%
The \textbf{AFEW-VA} dataset \cite{kossaifi2017afew} consists of 600 video clips (without audio) totalling to around 30K frames that were annotated in terms of valence-arousal; the values are discrete in the range of [-10,10]. The values have been scaled to [-1,1] so as to be consistent with the other datasets.

%%%%%%%%%%%%%%%%%%%%%%%%%%%%%%%%%%%%%%%%%%%%%%%%%%%%%%
%%%%%%%%%%%%%%%%%%%%%%%%%%%%%%%%%%%%%%%%%%%%%%%%%%%%%%
The \textbf{AffectNet} database \cite{mollahosseini2017affectnet} contains around 1M facial images, 400K of which were manually annotated in terms of 7 discrete expressions (plus contempt) and valence-arousal that ranges in $[-1,1]$.
The training set of this database consists of around 321K images and the validation of 5K. The validation set is balanced across the different emotion categories. 
%%%%%%%%%%%%%%%%%%%%%%%%%%%%%%%%%%%%%%%%%%%%%%%%%%%%%%
%%%%%%%%%%%%%%%%%%%%%%%%%%%%%%%%%%%%%%%%%%%%%%%%%%%%%%
The \textbf{RAF-DB} database \cite{li2017reliable} contains 12.2K training and 3K test facial images annotated in terms of the 7 basic and 11 compound emotion categories. 

%%%%%%%%%%%%%%%%%%%%%%%%%%%%%%%%%%%%%%%%%%%%%%%%%%%%%%
%%%%%%%%%%%%%%%%%%%%%%%%%%%%%%%%%%%%%%%%%%%%%%%%%%%%%%
The \textbf{EmotioNet} database \cite{fabian2016emotionet} contains around 1M images and was released for the EmotioNet Challenge in 2017 \cite{benitez2017emotionet}. 950K images were automatically annotated and the remaining 50K images were manually annotated with 11 AUs (1,2,4,5,6,9,12,17,20,25,26); around half of the latter constituted the validation and the other half the test set of the Challenge. Additionally, a subset of about 2.5K images was annotated with the 6 basic and 10 compound emotions.
%%%%%%%%%%%%%%%%%%%%%%%%%%%%%%%%%%%%%%%%%%%%%%%%%%%%%%
%%%%%%%%%%%%%%%%%%%%%%%%%%%%%%%%%%%%%%%%%%%%%%%%%%%%%%
The \textbf{DISFA} database \cite{mavadati2013disfa} is a lab controlled database consisting of 27 videos each of which has 4,845 frames, where each frame is coded with the AU intensity on a six-point discrete scale. AU intensities equal or greater than 2 are considered as occurrence, while others are treated as non-occurrence. There are in total 12 AUs (1,2,4,5,6,9,12,15,17,20,25,26). 
%%%%%%%%%%%%%%%%%%%%%%%%%%%%%%%%%%%%%%%%%%%%%%%%%%%%%%
%%%%%%%%%%%%%%%%%%%%%%%%%%%%%%%%%%%%%%%%%%%%%%%%%%%%%%
The \textbf{GFT} database consists of 96 videos of 96 subjects totalling around 130K frames. It is annotated for the occurrence of 10 AUs (1,2,4,6,10,12,14,15,23,24). The training set consists of 78 subjects of around 108K frames  and the test set of 18 subjects of around 24.5K frames.

%%%%%%%%%%%%%%%%%%%%%%%%%%%%%%%%%%%%%%%%%%%%%%%%%%%%%%
%%%%%%%%%%%%%%%%%%%%%%%%%%%%%%%%%%%%%%%%%%%%%%%%%%%%%%
The \textbf{BP4D-Spontaneous} database\cite{zhang14bp4d} (in the rest of the paper we refer to it as BP4D) contains 61 subjects with 223K frames and is annotated for the occurrence and intensity of 27 AUs. There are 21 subjects with 75.6K images in the training, 20 subjects with 71.2K images in the  development and 20 subjects with 75.7K images in the test partition. This database has been used as a part of the FERA 2015 Challenge \cite{valstar2015fera}, in which only 11 AUs (1,2,4,6,7,10,12,14,15,17,23) were used. 
%%%%%%%%%%%%%%%%%%%%%%%%%%%%%%%%%%%%%%%%%%%%%%%%%%%%%%
%%%%%%%%%%%%%%%%%%%%%%%%%%%%%%%%%%%%%%%%%%%%%%%%%%%%%%
The \textbf{BP4D+} database \cite{zhang2016multimodal} is an extension of BP4D incorporating different modalities as well as more subjects (140). It is annotated for occurrence of 34 AUs and intensity for 5 of them. It has been used as a part of the FERA 2017 Challenge \cite{valstar2017fera}, in which only 10 AUs (1,4,6,7,10,12,14,15,17,23) were used. There are 2952 videos of 41 subjects with 9 different views in the training set, 1431 of 20 subjects  with 9 different views in the validation set and 1080 videos of 30 subjects in the test set. 

Here let us note that for AffectNet, BP4D and BP4D+, no test set is released; thus we use the released validation set to test on and divide the training set into subject independent training and validation subsets.

In the face recognition case, we utilized the \textbf{CelebA} database. It contains 202,600 images of 10,177K identities (celebrities), each with 40 attribute annotations. It contains around 160K images in the  training, 20K in the validation and 20K in the test set. These sets are subject independent. For our experiment, we generated a new split into three subject dependent sets; each set contained images from all 10,177 identities. 

%\vspace*{-0.5cm}
\subsection{Performance Measures}
We use: \\
i) the CCC for Aff-Wild and Aff-Wild2 (CCC was the evaluation criterion of the respective Challenges), Affectnet and AFEW-VA, \\
%ii) the total accuracy for AFEW (this metric was the evaluation criterion of the EmotiW Challenges), 
ii) the mean diagonal value of the confusion matrix for RAF-DB (this criterion was selected for evaluating the performance on this database by \cite{li2017reliable}); the accuracy for AffectNet, \\
iii) the F1 score for DISFA, GFT, BP4D and BP4D+ (this metric was the evaluation criterion of the FERA 2015 and 2017 Challenges); for AU detection in EmotioNet the Challenge's metric was the average between: a) the mean (across all AUs) F1 score and b) the mean (across all AUs) accuracy; for the expression classification, it was the average between:  a) the average (across all emotions) F1 score and b) the unweighted average recall (UAR) over all emotion categories, \\
iv) the total accuracy and average F1 score for the attributes and ids in CelebA.

%\vspace*{-0.5cm}
\subsection{Pre- and Post-Processing}

\noindent \textit{Data Cleaning}: We performed data cleaning on the AffectNet database that contains overlapping annotations for valence-arousal and 7 basic expressions. We removed images for which there was a mismatch between the values of valence and
arousal and the discrete expressions. In more detail: i) images annotated as neutral should have radius of the valence-arousal vector smaller than 0.15; ii)  images annotated as sad or disgusted or fearful should have negative valence; iii) images annotated as angry should have negative valence and positive arousal; iv) images annotated as happy should have positive valence.

\smallskip

\noindent \textit{Data Subsampling}: Aff-Wild is a video database with consecutive frames having the same (or very similar) values for both valence and arousal. However FaceBehaviorNet is a CNN that does not exploit the temporal dependencies between frames; thus in each video, for each frame that we kept, we skipped the following four.

\smallskip

\noindent \textit{Face Pre-Processing}: We used the SSH detector \cite{najibi2017ssh} based on ResNet and trained on the WiderFace dataset \cite{yang2016wider} to extract, from all images, face bounding boxes and 5 facial landmarks; the latter were used  for face alignment. In the CelebA case we used the aligned data distributed with the database. All   cropped   and   aligned   images   were  resized   to $112 \times 112 \times 3$ pixel resolution and their intensity values were normalized  to  $[-1,1]$. 

\smallskip

\noindent \textit{Prediction Post-Processing}: Because Aff-Wild is a video database and FaceBehaviorNet is a CNN that does not exploit the temporal dependencies between frames, we performed median filtering of the - per frame - predictions.% with window size of 117.

\subsection{Training Implementation Details}\label{training}

At this point let us describe the strategy that was used, in the affect computing task, for feeding images from different databases to FaceBehaviorNet during training. At first, the training set was split into three different sets, each of which contained images that were annotated in terms of either valence-arousal, or action units, or seven basic expressions; let us denote these sets as VA-Set, AU-Set and EXPR-Set, respectively. During training, at each iteration, three batches, one from each of these sets (as can be seen in Fig.\ref{facebehaviornet}), were concatenated and fed to FaceBehaviorNet. This step is important for network training, because: i) the network minimizes the objective function of eq. \ref{eq:mt1}; at each iteration, the network has seen images from all categories and thus all loss terms contribute to the objective function, ii) since the network sees an adequate number of images from all categories, the weight updates (during gradient descent) are not based on noisy gradients; this in turn prevents poor convergence behaviors; otherwise, we would need to tackle these problems, e.g. do asynchronous SGD as proposed in \cite{kokkinos2017ubernet} to make the task parameter updates decoupled, 
iii) the CCC cost function (defined in Section \ref{approach}) needs an adequate sequence of predictions.

Since VA-Set, AU-Set and EXPR-Set had different sizes, they needed to be  'aligned'. To do so, we selected the batches of these sets in such a manner, so that after one epoch we will have sampled all images in the sets. In particular, we chose batches of size 200, 124 and 52 for the AU-Set, VA-Set and EXPR-Set, respectively. 
The training of FaceBehaviorNet was performed in an end-to-end manner, using the Momentum optimizer with 0.9 momentum and a learning rate of $10^{-4}$. Training was performed on a Tesla V100 32GB GPU; training time was about 2 days. The TensorFlow platform has been used.

\begin{table*}[ht]
\caption{Performance evaluation of valence-arousal, seven basic expression and action units predictions on all used databases provided by the FaceBehaviorNet when trained with/without the coupling losses, under the two task relatedness scenarios; 'AFA Score' is the average between the F1 Score and the Accuracy}
\label{comparison_losses}
\centering
\scalebox{0.95}{
\begin{tabular}{ |c||c||c|c|c|c|c|c|c|c|c|c| }
 \hline
\multicolumn{1}{|c||}{\begin{tabular}{@{}c@{}} Databases \end{tabular}} & \multicolumn{1}{c||}{\begin{tabular}{@{}c@{}} Relatedness \end{tabular}}  & \multicolumn{2}{c|}{Aff-Wild} & \multicolumn{3}{c|}{\begin{tabular}{@{}c@{}}  AffectNet \end{tabular}}   & \multicolumn{1}{c|}{RAF-DB} & \multicolumn{1}{c|}{EmotioNet} & \multicolumn{1}{c|}{DISFA}  & \multicolumn{1}{c|}{BP4D} & \multicolumn{1}{c|}{BP4D+}  \\
 \hline
 FaceBehaviorNet &   
 &\multicolumn{1}{c|}{CCC-V}  
 &\multicolumn{1}{c|}{CCC-A} 
 &\multicolumn{1}{c|}{CCC-V} 
 &\multicolumn{1}{c|}{CCC-A} &\multicolumn{1}{c|}{\begin{tabular}{@{}c@{}}  F1 \\ Score \end{tabular}}  
 & \multicolumn{1}{c|}{\begin{tabular}{@{}c@{}}  Mean diag. \\ of conf. matrix \end{tabular}} &\multicolumn{1}{c|}{\begin{tabular}{@{}c@{}}  AFA \\ Score \end{tabular}} 
 & \multicolumn{1}{c|}{\begin{tabular}{@{}c@{}}  F1 \\ Score \end{tabular}}
 &\multicolumn{1}{c|}{\begin{tabular}{@{}c@{}}  F1 \\ Score \end{tabular}} 
 &\multicolumn{1}{c|}{\begin{tabular}{@{}c@{}}  F1 \\ Score \end{tabular}} \\ 
  \hhline{=:=:=:=:=:=:=:=:=:=:=:=}
no coupling loss & -  & 0.52 & 0.38 & 0.56 & 0.50 & 0.59  & 0.67 & 0.72 & 0.54 & 0.76 & 0.56  \\
 \hline
 \hline
co-annotation & \cite{du2014compound} & 0.63  & 0.45  & 0.63  & 0.54 & 0.60   & 0.72  & 0.75  &  0.56 & 0.81 & 0.58  \\
 \hline
soft co-annotation & \cite{du2014compound}  & 0.64 & 0.46 & 0.64  & 0.55 & 0.62    & 0.74  & 0.76  & 0.57  & 0.82 & 0.60  \\
 \hline
 distr-matching & \cite{du2014compound}  & 0.64  & 0.44 & 0.64  & 0.57 & 0.62    & 0.75  &0.76   & 0.58 & 0.84  & 0.58  \\ 
 \hline 
\begin{tabular}{@{}c@{}} \textbf{soft co-annotation} \\ \textbf{and distr-matching}  \end{tabular} & \cite{du2014compound} & \textbf{0.67}   &  0.48   & \textbf{0.66} & \textbf{0.58}  & \textbf{0.65}      & 0.77  &  \textbf{0.77}   & 0.60  & \textbf{0.85}  & 0.60   \\
 \hline 
 \hline
co-annotation & Aff-Wild2 & 0.61  & 0.45  & 0.60  & 0.52 & 0.58  & 0.73  & 0.74 &  0.55 & 0.77 & 0.58  \\
\hline
soft co-annotation & Aff-Wild2  & 0.63 & 0.46 & 0.62  & 0.53 & 0.60   & 0.75  & 0.76  & 0.60  & 0.80 & 0.61  \\
 \hline
 distr-matching & Aff-Wild2  & 0.64  & 0.48 & 0.65  & 0.57 & 0.62   & 0.76  & 0.75 & 0.59 & 0.79  & 0.60  \\ 
 \hline 
\begin{tabular}{@{}c@{}} \textbf{soft co-annotation} \\ \textbf{and distr-matching}  \end{tabular} &  Aff-Wild2 &  \textbf{0.67}   & \textbf{0.49}   & \textbf{0.66} & 0.57  & \textbf{0.65}      & \textbf{0.78}  &  \textbf{0.77}   & \textbf{0.62}  & 0.83  & \textbf{0.62}  \\ 
 \hline
\end{tabular}
}
\end{table*}

\subsection{ Case Study I: Affective Computing}

\subsubsection{Results: Ablation Study}

First, we compared the performance of FaceBehaviorNet when trained with the losses of eq. \ref{eq:mt1} and: i) without using the coupling losses described in Section \ref{approach}, ii) with co-annotation coupling loss, iii) with soft co-annotation coupling loss, iv) with distr-matching coupling loss and v) with soft co-annotation and distr-matching coupling losses. Table \ref{comparison_losses} shows the results for all these approaches, when the task relatedness was drawn from the cognitive study, or when it was inferred empirically from Aff-Wild2.

Many deductions can be made. 
\textit{Firstly}, when FaceBehaviorNet was trained with a coupling loss, or with any combination of coupling losses, it displayed a better performance than when trained with no coupling loss. This holds on all databases and in both  task relatedness scenarios. This validates the fact that the proposed losses helped to couple the three studied tasks regardless of which relatedness scenario was followed.
\textit{Secondly}, the performance in estimation of valence and arousal was improved, although we did not explicitly designed a coupling loss for this; we only coupled emotion categories and action units. We conjecture that when action unit detection and expression classification accuracy  improves (due to coupling), valence and arousal performance also improves, because valence and arousal are implicitly coupled with emotions via joint dataset annotations.

% \textit{Thirdly}, in all scenarios, the co-annotation loss results in FaceBehaviorNet having the worst performance compared to all other coupling losses. \textcolor{red}{In any case, co-annotation loss was the basis for the development of the other two losses and thus it is expected to display the worst performance.}
\textit{Finally},  in both scenarios, across all databases, best results have been achieved when FaceBehaviorNet was trained with both soft co-annotation and distr-matching losses. In particular, in both settings, an average performance increase of more than 2.5\% has been observed when using both coupling losses, compared to the cases when only one of them was used. One can also observe that when task relatedness was inferred from Aff-Wild2, FaceBehaviorNet -trained with both coupling losses- displayed an average performance gain of 0.5\%  in expression classification and 0.5\% in AU detection compared to the case when task relatedness was inferred from the cognitive study of \cite{du2014compound}.

Next, we utilized three state-of-the-art and broadly used networks,  VGG-FACE \cite{parkhi2015deep}, ResNet-50 \cite{he2016deep} and DenseNet-121 \cite{huang2017densely}. We trained these networks in a multi-task manner (without using any coupling loss) with all the databases described in Section \ref{databases} and compared their performance to that of FaceBehaviorNet (trained without any coupling loss). As shown in Table \ref{comparison_nets}, the FaceBehaviorNet  has  proven  to  provide  the  best  results,  outperforming the MT-VGG-FACE by 1.6\%, MT-ResNet by 1.4\% and the MT-DenseNet by 4\%, on average (across all databases' metrics).

%%% maybe combine this table with tables 1,2,3 to gain space
\begin{table}[ht]
\caption{Performance comparison of different widely used networks and FaceBehaviorNet, when used in the multi-task framework (without coupling loss)}
\label{comparison_nets}
\centering
\scalebox{0.88}{
\begin{tabular}{ |c||c|c|c|c|c| }
 \hline
\multicolumn{1}{|c||}{Databases} & \multicolumn{1}{c|}{} & \multicolumn{1}{c|}{\begin{tabular}{@{}c@{}}  Face- \\ BehaviorNet \end{tabular}}  & \multicolumn{1}{c|}{\begin{tabular}{@{}c@{}}  MT- \\ VGG-FACE \end{tabular}} & \multicolumn{1}{c|}{\begin{tabular}{@{}c@{}}  MT- \\ ResNet \end{tabular}} & \multicolumn{1}{c|}{\begin{tabular}{@{}c@{}}  MT- \\ DenseNet \end{tabular}}  \\
  %\hhline{=:=:=:=:=}
  \hline \hline
Aff-Wild & \begin{tabular}{@{}c@{}}  CCC-V \\ CCC-A \end{tabular}  & \begin{tabular}{@{}c@{}} \textit{0.52} \\ \textbf{0.38} \end{tabular}   & \begin{tabular}{@{}c@{}}\textbf{ 0.55} \\ 0.36 \end{tabular}  & \begin{tabular}{@{}c@{}}  0.54 \\ \textbf{0.38} \end{tabular}   & \begin{tabular}{@{}c@{}}  0.52 \\ 0.35 \end{tabular}      \\
\hline
AffectNet & \begin{tabular}{@{}c@{}}  CCC-V \\ CCC-A \\ F1 Score \end{tabular} &  \begin{tabular}{@{}c@{}} \textbf{0.56} \\ \textbf{0.50} \\ \textbf{0.59} \end{tabular}     & \begin{tabular}{@{}c@{}}  \textbf{0.56} \\ 0.46 \\ 0.53 \end{tabular} & \begin{tabular}{@{}c@{}}  0.53 \\ 0.48 \\ 0.54 \end{tabular} & \begin{tabular}{@{}c@{}}  0.53 \\ 0.44 \\ 0.52 \end{tabular} \\
\hline
RAF-DB  & \begin{tabular}{@{}c@{}}  Mean \\ diag. of \\ conf. matrix \end{tabular}  &  \textbf{0.67}   & \textbf{0.67} & \textbf{0.67} & 0.64 \\
\hline
Emotionet  &\begin{tabular}{@{}c@{}}  AFA Score \end{tabular}   &    \textbf{0.72}   & \begin{tabular}{@{}c@{}} \textbf{0.72} \end{tabular} & \begin{tabular}{@{}c@{}}  0.71 \end{tabular} & \begin{tabular}{@{}c@{}}  0.69 \end{tabular} \\
\hline
DISFA & F1 score &    \textbf{0.54}     & 0.52 & 0.52 & 0.49 \\
\hline
BP4D & F1 score &  \textbf{0.76}    & 0.70 & 0.73 & 0.68 \\
\hline
BP4D+ & F1 score &    \textit{0.56}    & \textbf{0.57} & \textit{0.56} & 0.54 \\
\hline
\end{tabular}
}
\end{table}

\subsubsection{Results: Comparison with State-of-the-Art and Single-Task Methods}

\begin{table*}[ht]
\caption{Performance evaluation of valence-arousal, seven basic expression and action units predictions on all utilized databases provided by the FaceBehaviorNet and state-of-the-art methods; 'AFA Score' is the average between the F1 Score and the Accuracy; in parenthesis is the result in Aff-Wild without post-processing} %on Aff-Wild and AffectNet databases}
\label{comparison_sota}
\centering
\scalebox{0.95}{
\begin{tabular}{ |c||c|c|c|c|c|c|c|c|c|c| }
 \hline
\multicolumn{1}{|c||}{\begin{tabular}{@{}c@{}} Databases  \end{tabular}} & 
\multicolumn{2}{c|}{Aff-Wild} & \multicolumn{3}{c|}{\begin{tabular}{@{}c@{}}  AffectNet \end{tabular}}  & \multicolumn{1}{c|}{RAF-DB} & \multicolumn{1}{c|}{EmotioNet} & \multicolumn{1}{c|}{DISFA}  & \multicolumn{1}{c|}{BP4D} & \multicolumn{1}{c|}{BP4D+}   \\
 \hline
  & CCC-V & CCC-A & CCC-V & CCC-A &\begin{tabular}{@{}c@{}} Accuracy \end{tabular}  & \begin{tabular}{@{}c@{}}  Mean diagonal \\ of conf. matrix \end{tabular}  &\multicolumn{1}{c|}{\begin{tabular}{@{}c@{}}  AFA \\ Score \end{tabular}} &   \multicolumn{1}{c|}{\begin{tabular}{@{}c@{}}  F1 \\ Score \end{tabular}}&\multicolumn{1}{c|}{\begin{tabular}{@{}c@{}}  F1 \\ Score \end{tabular}} &\multicolumn{1}{c|}{\begin{tabular}{@{}c@{}}  F1 \\ Score \end{tabular}}   \\ 
  \hline
 \hline
best performing CNN\cite{kollias} \cite{kollias2018deep} & 0.51 & 0.33 & - & - & - & - & -  &  - &-  & -  \\
\hline 
AffWildNet\cite{kollias} \cite{kollias2018deep} & 0.57 & 0.43 & - & - & - & - & -  &  - &-  & -  \\
\hline 
(2 $\times$ ) AlexNet \cite{mollahosseini2017affectnet} & - & - & 0.60 & 0.34 & 0.58 & - & -   & - &-  & -   \\ 
\hline
VGG-FACE \cite{kollias2020va} & - & - & 0.61 & 0.48 & - & - & -   & - &-  & -   \\ 
\hline
RAN-ResNet18$^{+}$ \cite{wang2020region} & - & - & -   & - & 0.60 & - & - & - & -  & -  \\
\hline
VGG-FACE-mSVM\cite{li2017reliable}& - & - & -   & - & - & 0.58 & - & - & -  & -  \\
\hline 
DLP-CNN \cite{li2017reliable} & - & - & -   & - & - & 0.74 & - & - & -  & -  \\
\hline
AlexNet \cite{benitez2017emotionet} & - & -  &-  & - & - & - & 0.61 & - & - & -  \\ 
 \hline
ResNet-34 \cite{ding2017facial} & - & - & -  & - & - & - & 0.73 & - & -  & - \\ 
\hline
 JAA-Net \cite{shao2018deep} & - & -  &-  & - & - & - & - & 0.56  & - & -  \\
 \hline
 LP-Net \cite{niu2019local} & - & -  &-  & - & - & - & - & 0.57 & - & -  \\
 \hline
DLE extension \cite{yuce2015discriminant} & - & - & - &-  & - & - & -  & -   &0.51 & -  \\
\hline 
VGG-FACE \cite{tang2017view} & -  & - &-  & - & - & -  & -  & - & -  & 0.48  \\
\hline
ARL \cite{shao2019facial} & -  & - &-  & - & - & -  & -  & - & -  & 0.51  \\
\hline
\hline
(3 $\times$) ST-FaceBehaviorNet & \begin{tabular}{@{}c@{}} 0.60 \\ (0.57)  \end{tabular} & \begin{tabular}{@{}c@{}} 0.39 \\ (0.35) \end{tabular} & 0.55 & 0.48 & 0.60  & 0.71 & 0.69 & 0.52 & 0.61 & 0.48 \\
\hhline{=:=:=:=:=:=:=:=:=:=:=}
\begin{tabular}{@{}c@{}}  FaceBehaviorNet, no coupling loss \end{tabular} & \begin{tabular}{@{}c@{}} 0.56 \\ (0.52) \end{tabular} & \begin{tabular}{@{}c@{}} 0.42 \\ (0.38) \end{tabular} & 0.56 & \textit{0.50} & 0.59  & 0.67 & 0.72 & 0.54 & \textit{0.76} & \textit{0.56}  \\
\hline
\begin{tabular}{@{}c@{}} \textbf{FaceBehaviorNet, soft co-annotation} \\ \textbf{and distr-matching, \cite{du2014compound}}  \end{tabular} &  \begin{tabular}{@{}c@{}} \textbf{0.67} \\ \textbf{(0.63)} \end{tabular}  & \begin{tabular}{@{}c@{}} \textit{0.48}  \\ \textit{(0.44)} \end{tabular}  & \textbf{0.66} & \textbf{0.58}  & \textbf{0.65}      & \textit{0.77}  &  \textbf{0.77}   & \textit{0.60}  & \textbf{0.85}  & \textit{0.60}  \\ 
\hline
\begin{tabular}{@{}c@{}} \textbf{FaceBehaviorNet, soft co-annotation} \\ \textbf{and distr-matching, Aff-Wild2}  \end{tabular} & \begin{tabular}{@{}c@{}} \textbf{0.67} \\ \textbf{(0.63)} \end{tabular}  & \begin{tabular}{@{}c@{}} \textbf{0.49}  \\ \textbf{(0.45)} \end{tabular}  & \textbf{0.66} & \textit{0.57}  & \textbf{0.65}      & \textbf{0.78}  &  \textbf{0.77}   & \textbf{0.62}  & \textit{0.83}  & \textbf{0.62}  \\ 
\hline
\end{tabular}
}
\end{table*}

Next, we trained a single-task FaceBehaviorNet (ST-FaceBehaviorNet) on all  dimensionally annotated databases, so as to predict valence and arousal; we also trained another similar network on all categorically annotated databases, to perform seven basic expression classification; finally we trained a third similar network on all databases annotated with action units, so as to perform AU detection. All these networks were based on residual units and had the same structure as FaceBehaviorNet; their only difference was the output layer. For brevity, these three single-task networks are denoted as '(3 $\times$) ST-FaceBehaviorNet' in one row of Table \ref{comparison_sota}.

We compared these networks' performance with the performance of FaceBehaviorNet when trained with and without the coupling losses. We also compared them with the performance of the state-of-the-art (sota) methodologies for each utilized database: 
i) the best performing CNN (VGG-FACE)  \cite{kollias}\cite{kollias2018deep} on Aff-Wild; 
ii) the best performing network (AffWildNet)  \cite{kollias}\cite{kollias2018deep} on Aff-Wild; 
iii) the baseline networks (AlexNet)  \cite{mollahosseini2017affectnet} on AffectNet  (in Table \ref{comparison_sota} they are denoted as '(2 $\times$) AlexNet' as they are two different networks: one for VA estimation and another for expression classification); 
iv) the state-of-the-art VGG-FACE \cite{kollias2020va} for VA estimation on AffectNet;
v) the state-of-the-art RAN-ResNet18$^{+}$ \cite{wang2020region} for expression classification on AffectNet;
vi) the VGG-FACE-mSVM \cite{li2017reliable} on RAF-DB; 
vii) the best performing network (DLP-CNN) \cite{li2017reliable} on RAF-DB; 
viii) the baseline network (AlexNet) \cite{benitez2017emotionet} on EmotioNet ;
ix) the winner of EmotioNet Challenge and best performing network (ResNet-34) \cite{ding2017facial} on EmotioNet;
x) the state-of-the-art network (LP-Net) \cite{niu2019local} on DISFA;
xi) the best performing network (LP-Net) \cite{niu2019local} on DISFA;
xii) the winner of FERA 2015, DLE extension \cite{yuce2015discriminant} on BP4D;
xiii) the winner of FERA 2017 (VGG-FACE) \cite{tang2017view} on BP4D+;
xiv) the best performing network (ARL) \cite{shao2019facial} on BP4D+. 
Table \ref{comparison_sota} displays the performance of all these networks.

\smallskip

\noindent \underline{\textit{Non-coupled MTL vs Sota \& ST}}: In Table \ref{comparison_sota} it can be seen that when no coupling loss is used for training FaceBehaviorNet, the network outperforms the state-of-the-art in BP4D+ by 5\%, BP4D by 25\% for AU detection and in AffectNet by 2\% for arousal estimation. In all other databases it displays either a slightly worse performance (1\% in three databases), or worse performance (3\%, 5\% and 7\% in three other databases).
Table \ref{comparison_sota} additionally illustrates a comparison between the three single task ST-FaceBehaviorNet and (multi-task) FaceBehaviorNet trained without any coupling loss. 
It can be observed that the multi-task network displays a better performance for AU detection and VA estimation, but an inferior one for expression classification. The latter indicates that negative transfer occurs in the case of basic expressions.

%\textcolor{red}{It is also worth mentioning that maybe the negative transfer in expressions was a bit transferred to VA (as expressions and VA are implicitly coupled via joint annotations in AffectNet). Although, overall, for VA estimation, the multi-task model outperformed the single task one (for arousal in Aff-Wild and valence-arousal estimation in AffectNet the multi-task network showed a better performance), for valence estimation in Aff-Wild the multi-task network showed an inferior performance to the single task one; therefore that is why we state that maybe the negative transfer in expressions was transferred to VA.}

This negative transfer effect was caused by the fact that some of the related tasks - valence-arousal and action units - dominated the training process. It should be mentioned that these two tasks included more data, i.e., images and corresponding annotations,  than the expression recognition task. In particular, there were twice more data in VA and three times more data in AUs compared to basic expressions. Negative transfer largely depends on the size of labeled data per task \cite{wang2019characterizing}. In fact, the amount of labeled data per task has  a direct effect on the feasibility and reliability of discovering shared regularities between the joint distributions of the tasks in MTL.  

A way of overcoming negative transfer in expression recognition would be to change (i.e., increase) the lambda value in the expression loss (that controls the relative importance of the task), or decrease the lambda values in the VA and AU losses in the total loss function of eq. \ref{eq:mt1} that is minimized by the multi-task model during training. However, this could severely affect the performance of the other tasks. Furthermore, this lambda hyper-parameter tuning is a computationally expensive procedure, which lasts many days for each trial. It should be added that this is an ad-hoc methodology which does not guarantee to work on other tasks, or in other databases. In order to balance the performance on many tasks, \cite{liu2019loss} proposed an iterative method which uses the training loss of each task to indicate whether it is well trained or not, and then decreases the relative weights of the well trained tasks. The problem with this approach is that it is based on costly evaluation of performance indicators during each training iteration and that this is performed in a rather task-agnostic way.

Negative transfer may be induced by conflicting gradients among the different tasks \cite{yu2020gradient}. Searching for Pareto solutions \cite{lin2019pareto} could remedy this. 
\cite{lin2019pareto} tackled  multi-task learning problems through multi-objective optimization, with decomposition  of the problem into a set of constrained sub-problems with different trade-off preferences (among different tasks). However, this approach is rather complex, providing a finite set of solutions that do not always satisfy the MTL requirements and finally need to perform trade-offs among tasks.

Through the proposed coupling loss, knowledge of the task relationship was infused in network training, thus providing it, in a simple manner, with higher level representation of the relationship between the tasks;  it was not based on performance indicators and it did not perform any trade-offs between the different tasks. The fact that the proposed coupling losses tackled the negative transfer is illustrated by the performance shown in Tables \ref{comparison_sota} and \ref{comparison_losses}, where FaceBehaviorNet trained with the proposed coupling losses outperformed by a large margin both the independently trained single task networks and the multi-task network trained without any coupling loss.

\smallskip

\noindent \underline{\textit{Coupled MTL vs Non-coupled MTL, ST \& Sota}}: In Table \ref{comparison_sota}, it can be observed that FaceBehaviorNet, when trained with the two coupling losses (in either task relatedness setting), outperformed FaceBehaviorNet trained without any coupling loss by: 9\% (average CCC) on Aff-Wild; 8.5\% (average CCC) and 6\% (accuracy) on AffectNet; 11\% on RAF-DB; 5\% on EmotioNet; 8\% on DISFA; 7\% on BP4D and 6\% on BP4D+. It further outperformed by a large margin the three ST-FaceBehaviorNet networks (10.3\% on average across all databases, with the minimum difference in performance being 5\% and the maximum 22\%).
Finally, in Table \ref{comparison_sota}, it can be observed that FaceBehaviorNet, when trained with the two coupling losses (in either task relatedness setting), outperformed by a very large margin \textit{every} state-of-the-art method in \textit{all} databases in \textit{all} tasks. More precisely, it outperformed:
\begin{itemize}[topsep=1pt,itemsep=0pt,partopsep=1pt, parsep=1pt,leftmargin=*]
    \item[i.] AffWildNet by 8\% (average CCC) for VA estimation on Aff-Wild; this is despite the fact that AffWildNet is a CNN-RNN that exploited the fact that Aff-Wild is an audio-visual database and despite the fact that facial landmarks were provided as additional inputs to the network, thus improving its performance; it also outperformed the best performing CNN in Aff-Wild by 16\%
    \item[ii.] VGG-FACE by 7.5\% (average CCC) for VA estimation on AffectNet, although VGG-FACE was trained with the original images plus thousands of generated images by VA-StarGAN; it also outperformed the AlexNet baseline by 15\% 
    \item[iii.] RAN-ResNet18$^{+}$ by 5\% for expression classification on AffectNet, despite the fact that the network was trained with a region based loss that encouraged a high attention weight for the most important regions in the input images; it also outperformed AlexNet (that used a weighted loss that heavily penalized the network for misclassifying examples from under-represented classes and penalized the network less for misclassifying examples from well-represented classes), by 7\%
    \item[iv.] DLP-CNN by 4\% for expression classification on RAF-DB, although this network was trained using a joint classical softmax loss - which forced different classes to stay apart - and a newly created loss - that pulled the locally neighboring faces of the same class together; it also outperformed VGG-FACE-mSVM, that used the standard cross entropy loss, by 20\%
    \item[v.] ResNet-34 by 4\% for AU detection on EmotioNet; it also outperformed AlexNet by 16\%
    \item[vi.] both JAA-Net and LP-Net by 6\% and 5\%, respectively, for AU detection on DISFA, despite the fact that these networks have additionally used facial landmarks as additional inputs, thus improving their performance
    \item[vii.] DLE extension by 34\% for AU detection on BP4D
    \item[viii.] ARL by 11\% for AU detection on BP4D+; it also outperformed VGG-FACE by 14\%
\end{itemize} 

At this point let us mention that in our approach the loss functions that we utilized (for expression classification and for AU detection) were standard losses (binary and softmax cross entropy, respectively). As was shown above, in most state-of-the-art approaches, more elaborate and advanced loss functions have been used. We could utilize more elaborate and advanced losses as well, but the focus of this work has been on the coupling between the tasks exploiting their relationship - inferred either empirically from external dataset annotations, or from a cognitive study; therefore we chose  not to use such losses whose specific selection could affect the presented analysis and the obtained results. As a future work, we will incorporate more advanced losses in the multi-task learning setting, which will further enhance the achieved results.

It might be argued that the more data used for network training (i.e., the additional data coming from the multiple tasks that are being solved, even if they contain partial or non-overlapping annotations), the better network performance will be in all tasks. However, as was shown, the three studied tasks are heterogeneous and negative transfer can occur, or sub-optimal models can be produced for some, or even all tasks \cite{wu2019understanding}.
As shown in Table \ref{comparison_sota}, FaceBehaviorNet, when trained with both proposed coupling losses, achieved a better performance on all databases than the independently trained single-task models. This illustrates that simultaneously 
%all described facial behavior tasks  are coherently correlated to each other; 
training this end-to-end architecture with the heterogeneous databases and coupling the corresponding tasks and annotations,  led to improved performance.
The fact that the network additionally outperformed the state-of-the-art methods, in both task relatedness settings, verified the generality of the proposed losses; network performance was boosted, independently of the type of task relatedness that was used.

\subsubsection{Results: Generalisation to Unseen Databases}

%isws vale ena section pou perigrafeis ta provlimata pou kanoume tackle:
%- more data => better performance : negative transfer
%- ambiguous cases in tasks k add incosistent au labelling: pes gia uncorrelated aus k isws bp4d au6 k au12 p einai correlated gia diafora tasks
%- pes gia bias tis kathe vasis
%- partial k non-overlapping annotations

At this point let us mention that each task and corresponding database contains ambiguous cases: i) there is generally discrepancy in the perception of the disgust, fear, sadness and (negative) surprise emotions across different people (of different ethnicity, race and age) and across databases; emotional displays and their perception are not universal, i.e.,  facial expressions are displayed and interpreted differently depending on the cultural background of subjects and annotators \cite{gendron2018universality,bryant2019comparative}
; ii) the exact valence and arousal value for a particular affect is also not consistent among databases; iii) the AU annotation process is hard to do and error prone, creating incosistency among databases (e.g., regarding the dependencies among AUs, such as  AU12 -lip corner puller- and AU15 -lip corner depressor- that cannot co-occur as their corresponding muscle groups -Zgomaticus major and Depressor anguli oris, respectively- are unlikely to be  simultaneously activated).   

Furthermore each database contains its own bias. The bias is either in terms of the race, ethnicity and ages of the subjects displayed in the database or the race, ethnicity, age and culture of the experts that performed the annotations (having their own bias in judging the depicted affect). When an affect recognition system is trained on one database, then the system inherently learns the bias towards facial displays present in the training data. Therefore, when the system is tested on another, new and unseen database (with other demographics and statistics), its performance is not as good. An example is the AffectNet database that contains images of European Americans and has been annotated mainly by European Americans. Another example is GFT database that mainly contains videos of white and black people.

The excellent generalization performance of FaceBehaviorNet -across the test sets- of the 7 databases whose training sets have been used for its training, is an indicator that the network and the proposed coupling losses tackled the aforementioned issues. By jointly training on all databases and by coupling the tasks, we overcame these limitations as shown in the extensive experimental study across the 7 databases. 
In  order  to  further illustrate and validate  that  FaceBehaviorNet learned good and robust features, we show that it is capable of generalizing its knowledge and capabilities in other new and unseen affect recognition databases that have not been utilized during its training and contain different statistics and contexts.

Table \ref{comparison_cross_db} compares the performance of FaceBehaviorNet when trained with both coupling losses with the performance of  state-of-the-art networks on two new databases. It is worth mentioning that these two databases, AFEW-VA and GFT, have not been utilized during FaceBehaviorNet's training and no fine-tuning was performed; FaceBehaviorNet was just tested on the new and unseen databases. It can be observed that FaceBehaviorNet outperformed both AffWildNet (that was further trained on AFEW-VA) and JAA-Net (that was trained on GFT) by 4\% (on average) and 5\%, respectively.

\begin{table}[ht]
\caption{Performance comparison of FaceBehaviorNet and state-of-the-art methods on two databases that were not utilized for its training; in parenthesis is the result without post-processing}
\label{comparison_cross_db}
\centering
\scalebox{1.}{
\begin{tabular}{ |c||c|c|c| }
 \hline
\multicolumn{1}{|c||}{\begin{tabular}{@{}c@{}} Databases \end{tabular}} &  \multicolumn{2}{c|}{AFEW-VA} & \multicolumn{1}{c|}{GFT}  \\
 \hline
 Methods  &\multicolumn{1}{c|}{\begin{tabular}{@{}c@{}} CCC-V \end{tabular}} &  \multicolumn{1}{c|}{\begin{tabular}{@{}c@{}} CCC-A \end{tabular}} &\multicolumn{1}{c|}{\begin{tabular}{@{}c@{}} F1 \\ Score \end{tabular}}   \\
  \hhline{=:=:=:=}
JAA-Net \cite{shao2018deep}  & - & - & 0.55 \\
\hline
fine-tuned AffWildNet\cite{kollias} \cite{kollias2018deep}  & 0.52 & 0.56 & - \\
\hline
\hline
\begin{tabular}{@{}c@{}} \textbf{FaceBehaviorNet both coupling losses}  \end{tabular}   &  \begin{tabular}{@{}c@{}} \textbf{0.57} \\ \textbf{(0.55)} \end{tabular} & \begin{tabular}{@{}c@{}}\textbf{0.59} \\ (0.57) \end{tabular}   & \textbf{0.60}  \\
\hline
\end{tabular}
}
\end{table}

\subsubsection{Results: Zero-Shot and Few-Shot Learning}

\noindent In order to further show and validate that FaceBehaviorNet learned good features encapsulating all aspects of facial behavior, we conducted zero- and few-shot learning experiments for classifying compound expressions.   
Given that there exist only 2 datasets (EmotioNet and RAF-DB) annotated with compound expressions and that they do not contain a lot of samples (less than 5,000 each), at first, we used the predictions of FaceBehaviorNet together with the rules from \cite{du2014compound} to generate compound emotion predictions in a zero-shot learning manner, as was described in Section \ref{zero-shot}. Additionally, to  demonstrate  the  superiority  of FaceBehaviorNet, we used it as a pre-trained network in a few-shot learning experiment. We took advantage of the fact that our network has learned good features and used them as priors for fine-tuning the network to perform compound emotion classification.

\smallskip

\noindent {\underline{RAF-DB database}}
At first, we performed zero-shot experiments on the 11 compound categories of RAF-DB. We computed a candidate score,  $\mathcal{C}_{s}(y_{emo})$, for each class $y_{emo}$ as shown in Section \ref{zero-shot}.
Table \ref{comparison_zero_few_shot} shows the results of this approach when we used the predictions of FaceBehaviorNet trained with and without the soft co-annotation and distr-matching losses. Best results have been obtained when the network was trained with the coupling losses. One can observe, that this approach outperformed by 15.1\% the baseline VGG-FACE-mSVM \cite{li2017reliable} that been trained on RAF-DB for compound emotion classification. It also outperformed by 2.1\% the state-of-the-art and best performing network, DLP-CNN that has also been trained on RAF-DB for compound emotion classification; DLP-CNN used a loss function designed for this specific task.

Next, we targeted few-shot learning. In particular, we fine-tuned FaceBehaviorNet (trained with and without the soft co-annotation and distr-matching losses) on the small training set of RAF-DB. In Table \ref{comparison_zero_few_shot}, it can be seen that the fine-tuned FaceBehaviorNet, trained with and without the coupling losses, outperformed by large margins, 23.7\% and 10.7\%, respectively, the baseline VGG-FACE-mSVM and the state-of-the-art and best performing DLP-CNN.

\smallskip

\noindent {\underline{EmotioNet database}}
Next, we performed zero-shot experiments on the EmotioNet basic and compound set that was released for the related Challenge. This set includes 6 basic plus 10 compound categories, as described at the beginning of this Section. Our zero-shot methodology was similar to the one described above for the RAF-DB database. 

The results of this experiment are shown in Table \ref{comparison_zero_few_shot}. 
Best results were also obtained when the network was trained with the two coupling losses. It can be observed that this approach outperformed by 5.7\% and 8.6\% in F1 score and Unweighted Average Recall (UAR), respectively, the state-of-the-art and winner of EmotioNet Challenge, NTechLab's \cite{benitez2017emotionet} approach, which used the Emotionet's images with compound annotation.

\begin{table}[ht]
\caption{Performance evaluation of generated compound emotion predictions; 'UAR' denotes Unweighted Average Recall
}
\label{comparison_zero_few_shot}
\centering
\scalebox{0.83}{
\begin{tabular}{ |c||c|c|c| }
 \hline
\multicolumn{1}{|c||}{\begin{tabular}{@{}c@{}} Databases \end{tabular}} & \multicolumn{2}{c|}{EmotioNet} & \multicolumn{1}{c|}{RAF-DB}  \\
 \hline
 Methods &\multicolumn{1}{c|}{\begin{tabular}{@{}c@{}}  F1 \\ Score \end{tabular}} &  \multicolumn{1}{c|}{\begin{tabular}{@{}c@{}} UAR \end{tabular}} &\multicolumn{1}{c|}{\begin{tabular}{@{}c@{}}  Mean diag. of \\ conf. matrix \end{tabular}}   \\
  \hhline{=:=:=:=}
\begin{tabular}{@{}c@{}}  zero-shot, FaceBehaviorNet, no coupling loss \end{tabular} & 0.243 & 0.260  &  0.382   \\
\hline
\begin{tabular}{@{}c@{}} \textbf{zero-shot,  FaceBehaviorNet, both coupling losses}  \end{tabular} & \textbf{0.312}  & \textbf{0.329}   &  \textbf{0.467}   \\
\hline
\hline
NTechLab \cite{benitez2017emotionet} & 0.255 & 0.243 & - \\
\hline
VGG-FACE-mSVM \cite{li2017reliable} & - & - & 0.316 \\
\hline
DLP-CNN \cite{li2017reliable} & - & - & 0.446 \\
\hline
\hline
\begin{tabular}{@{}c@{}} fine-tuned FaceBehaviorNet, no coupling loss \end{tabular} & - & -  &  0.468   \\
\hline
\begin{tabular}{@{}c@{}} \textbf{fine-tuned FaceBehaviorNet, both coupling losses} \end{tabular} & - & -  & \textbf{0.553}    \\
\hline
\end{tabular}
}
\end{table}

\subsection{Case Study II: Face Recognition}

At first, we trained a single-task FaceBehaviorNet (ST-FaceBehaviorNet) on CelebA to detect the 40 facial attributes; we also trained another similar network on CelebA to perform classification into the 10,177 different identities. The networks were based on residual units and had the same structure as FaceBehaviorNet; their only difference was the output layer. For brevity, these two single-task networks are denoted as '(2 $\times$) ST-FaceBehaviorNet' in one row of Table \ref{comparison_celeba}.
We compared these networks' performance with the performance of FaceBehaviorNet when trained with the distribution matching coupling loss and when trained without the coupling loss. 

Table \ref{comparison_celeba} shows that the multi-task FaceBehaviorNet, when trained without the coupling loss, outperformed the 2 single-task networks ST-FaceBehaviorNet in both studied tasks and in both metrics (Total Accuracy and average F1 Score). In more detail, it displayed an improved performance by 2.28\% and 2.1\% in the Accuracy and F1 Score metrics, respectively, for identity classification; it also displayed an improved performance by 1.91\% and 0.89\% in the Accuracy and F1 Score metrics, respectively, for attribute detection. This shows that the two studied facial heterogeneous tasks were coherently correlated to each other; training the end-to-end multi-task architecture therefore, led to improved performance and no negative transfer occurred. 

Table \ref{comparison_celeba} further shows that the multi-task FaceBehaviorNet, when trained with the distribution matching coupling loss, greatly outperformed its counterpart that was trained without that loss, in both studied tasks and in both metrics. More precisely, when training with the coupling loss, performance increased by 4.57\% and 5.9\% in the Accuracy and F1 Score metrics, respectively, for identity classification; performance also increased by 1.24\% and 2.3\% in the Accuracy and F1 Score metrics, respectively, for attribute detection. This proves the effectiveness of the proposed distribution matching loss.

Next, we utilized, as an ablation study, three state-of-the-art and widely used networks, VGG-FACE, ResNet-50 and DenseNet-121. We trained these networks in a multi-task manner (without using the coupling loss) on CelebA and compared their performance to that of FaceBehaviorNet that was also trained without coupling loss (for a fair comparison). As shown in Table \ref{comparison_celeba}, FaceBehaviorNet has proven to be the best architecture as it  provided  the  best  results,  outperforming on all tasks and metrics the MT-VGG-FACE by at least 2\%, the MT-ResNet by at least 1.8\% and the MT-DenseNet by at least 4\%.

Finally, let us mention that the achieved performance of FaceBehaviorNet trained with the distribution matching loss for attribute detection was higher than the performance of various state-of-the-art methodologies on the same database. FaceBehaviorNet reached an accuracy of 93.22\%, whereas DMTL \cite{han2017heterogeneous} achieved 92.6\% accuracy, MCNN-AUX \cite{hand2016attributes} achieved an accuracy of 91.29\% and Face-SSD \cite{jang2019registration} achieved an accuracy of 90.29\%. These comparisons are provided just as an indication; they are not direct comparisons.

\begin{table}[ht]
\caption{Performance evaluation on CelebA; 'Acc' denotes the total accuracy}
\label{comparison_celeba}
\centering
\scalebox{1.}{
\begin{tabular}{ |c||c|c|c|c| }
 \hline
\multicolumn{1}{|c||}{\begin{tabular}{@{}c@{}} CelebA \end{tabular}} & \multicolumn{2}{c|}{Ids} & \multicolumn{2}{c|}{Attributes}   \\
 \hline
  &
 \multicolumn{1}{c|}{\begin{tabular}{@{}c@{}}  Acc \end{tabular}} & 
 \multicolumn{1}{c|}{\begin{tabular}{@{}c@{}} F1 \end{tabular}} &
 \multicolumn{1}{c|}{\begin{tabular}{@{}c@{}}  Acc \end{tabular}} &
 \multicolumn{1}{c|}{F1} \\
  \hline \hline
\begin{tabular}{@{}c@{}} MT-VGG-FACE \end{tabular} & 80.07  & 72.2  & 89.87  & 70.24  \\
\hline
\begin{tabular}{@{}c@{}} MT-ResNet \end{tabular} & 80.83  & 72.9  & 90.11  & 71.38  \\
\hline
\begin{tabular}{@{}c@{}} MT-DenseNet \end{tabular}  & 78.11  & 70.02  & 87.57  & 67.88  \\
\hline
\hline
\begin{tabular}{@{}c@{}} (2 $\times$)  ST-FaceBehaviorNet \end{tabular} & 80.43  & 73.5  & 90.07  & 71.78  \\
\hline
\hline
\begin{tabular}{@{}c@{}} FaceBehaviorNet, no coupling loss \end{tabular} & 82.71  & 75.6  & 91.98  & 72.67  \\
\hline
\begin{tabular}{@{}c@{}} \textbf{FaceBehaviorNet, distr-matching}\end{tabular} & \textbf{87.28}  & \textbf{81.5}  & \textbf{93.22}  & \textbf{74.97}  \\

\hline
\end{tabular}
}
\end{table}

%\parbox[t]{2mm}{\multirow{3}{*}{\rotatebox[origin=c]{90}{rota}}}

%+ table for all 40 attributes

\section{Conclusions}
In this paper, we target heterogeneous MTL, i.e., simultaneously addressing detection, classification and regression problems. We propose co-training task in a weakly-supervised way, by exploring their relatedness. Relatedness between  the heterogeneous  tasks  is  either  provided  explicitly  in  a form  of  expert  knowledge,  or  is  inferred  based  on  empirical studies. In  co-training,  the  related  tasks  exchange their predictions and iteratively teach each other so that predictors of  all  tasks  can  excel even  if  we  have  limited  or  no  data for some  of  them. We  propose  an  effective  distribution matching  approach  based  on  distillation,  where  knowledge exchange between tasks is enabled via distribution matching over their  predictions. Based  on  this  approach,  we  build  the  first holistic framework for large-scale face analysis, FaceBehaviorNet, with case studies in affective computing and in face recognition.

In the first case study, FaceBehaviorNet is trained for joint basic expression recognition, action unit detection and valence-arousal estimation. All publicly available databases that study facial behavior tasks in-the-wild, have been utilized. In the latter case study, FaceBehaviorNet is trained for joint facial attribute detection and face identification. An extensive experimental study -across 10 databases- is performed that compares the performance of the holistic (multi-task, multi-domain, multi-label) FaceBehaviorNet (trained with and without taking into account the task-relatedness) to the performance of the single-task networks, as well as to the performance of the state-of-the-art. FaceBehaviorNet consistently outperformed, by a large margin, all of them ,in all databases, in both case studies, even mitigating bias and tackling negative transfer.
Finally, we explored the feature representation learned by FaceBehaviorNet in the joint training and showed its generalization abilities on the task of compound expressions, under zero-shot and few-shot learning settings.

\ifCLASSOPTIONcaptionsoff
  \newpage
\fi

\bibliographystyle{IEEEtran}
\bibliography{egbib}

\begin{thebibliography}{10}
\providecommand{\url}[1]{#1}
\csname url@samestyle\endcsname
\providecommand{\newblock}{\relax}
\providecommand{\bibinfo}[2]{#2}
\providecommand{\BIBentrySTDinterwordspacing}{\spaceskip=0pt\relax}
\providecommand{\BIBentryALTinterwordstretchfactor}{4}
\providecommand{\BIBentryALTinterwordspacing}{\spaceskip=\fontdimen2\font plus
\BIBentryALTinterwordstretchfactor\fontdimen3\font minus
  \fontdimen4\font\relax}
\providecommand{\BIBforeignlanguage}[2]{{%
\expandafter\ifx\csname l@#1\endcsname\relax
\typeout{** WARNING: IEEEtran.bst: No hyphenation pattern has been}%
\typeout{** loaded for the language `#1'. Using the pattern for}%
\typeout{** the default language instead.}%
\else
\language=\csname l@#1\endcsname
\fi
#2}}
\providecommand{\BIBdecl}{\relax}
\BIBdecl

\bibitem{wang2015holistic}
S.~Wang, S.~Fidler, and R.~Urtasun, ``Holistic 3d scene understanding from a
  single geo-tagged image,'' in \emph{Proceedings of the IEEE Conference on
  Computer Vision and Pattern Recognition}, 2015, pp. 3964--3972.

\bibitem{ranjan2017all}
R.~Ranjan, S.~Sankaranarayanan, C.~D. Castillo, and R.~Chellappa, ``An
  all-in-one convolutional neural network for face analysis,'' in \emph{2017
  12th IEEE International Conference on Automatic Face \& Gesture Recognition
  (FG 2017)}.\hskip 1em plus 0.5em minus 0.4em\relax IEEE, 2017, pp. 17--24.

\bibitem{kokkinos2017ubernet}
I.~Kokkinos, ``Ubernet: Training a universal convolutional neural network for
  low-, mid-, and high-level vision using diverse datasets and limited
  memory,'' in \emph{Proceedings of the IEEE Conference on Computer Vision and
  Pattern Recognition}, 2017, pp. 6129--6138.

\bibitem{zamir2018taskonomy}
A.~R. Zamir, A.~Sax, W.~Shen, L.~J. Guibas, J.~Malik, and S.~Savarese,
  ``Taskonomy: Disentangling task transfer learning,'' in \emph{Proceedings of
  the IEEE Conference on Computer Vision and Pattern Recognition}, 2018, pp.
  3712--3722.

\bibitem{zhang2021survey}
Y.~Zhang and Q.~Yang, ``A survey on multi-task learning,'' \emph{IEEE
  Transactions on Knowledge and Data Engineering}, 2021.

\bibitem{pan2010survey}
S.~J. Pan and Q.~Yang, ``A survey on transfer learning,'' \emph{IEEE
  Transactions on knowledge and data engineering}, vol.~22, no.~10, pp.
  1345--1359, 2010.

\bibitem{hinton2015distilling}
G.~Hinton, O.~Vinyals, and J.~Dean, ``Distilling the knowledge in a neural
  network,'' \emph{arXiv:1503.02531}, 2015.

\bibitem{ekman1997face}
R.~Ekman, \emph{What the face reveals: Basic and applied studies of spontaneous
  expression using the Facial Action Coding System (FACS)}.\hskip 1em plus
  0.5em minus 0.4em\relax Oxford University Press, USA, 1997.

\bibitem{kollias2018deep}
D.~Kollias, P.~Tzirakis, M.~A. Nicolaou, A.~Papaioannou, G.~Zhao, B.~Schuller,
  I.~Kotsia, and S.~Zafeiriou, ``Deep affect prediction in-the-wild: Aff-wild
  database and challenge, deep architectures, and beyond,'' \emph{International
  Journal of Computer Vision}, feb 2019.

\bibitem{zafeiriou2017aff}
S.~Zafeiriou, D.~Kollias, M.~A. Nicolaou, A.~Papaioannou, G.~Zhao, and
  I.~Kotsia, ``Aff-wild: Valence and arousal ‘in-the-wild’challenge,'' in
  \emph{Computer Vision and Pattern Recognition Workshops (CVPRW), 2017 IEEE
  Conference on}.\hskip 1em plus 0.5em minus 0.4em\relax IEEE, 2017, pp.
  1980--1987.

\bibitem{mollahosseini2017affectnet}
A.~Mollahosseini, B.~Hasani, and M.~H. Mahoor, ``Affectnet: A database for
  facial expression, valence, and arousal computing in the wild,'' \emph{arXiv
  preprint arXiv:1708.03985}, 2017.

\bibitem{emotionet2016}
C.~Benitez-Quiroz, R.~Srinivasan, and A.~Martinez, ``Emotionet: An accurate,
  real-time algorithm for the automatic annotation of a million facial
  expressions in the wild,'' in \emph{Proceedings of IEEE International
  Conference on Computer Vision \& Pattern Recognition (CVPR'16)}, Las Vegas,
  NV, USA, June 2016.

\bibitem{openface2015}
T.~Baltru{\v{s}}aitis, M.~Mahmoud, and P.~Robinson, ``Cross-dataset learning
  and person-specific normalisation for automatic action unit detection,'' in
  \emph{2015 11th IEEE International Conference and Workshops on Automatic Face
  and Gesture Recognition (FG)}, vol.~6, 2015, pp. 1--6.

\bibitem{kollias2019expression}
D.~Kollias and S.~Zafeiriou, ``Expression, affect, action unit recognition:
  Aff-wild2, multi-task learning and arcface,'' \emph{arXiv preprint
  arXiv:1910.04855}, 2019.

\bibitem{du2014compound}
S.~Du, Y.~Tao, and A.~M. Martinez, ``Compound facial expressions of emotion,''
  \emph{Proceedings of the National Academy of Sciences}, vol. 111, no.~15, pp.
  E1454--E1462, 2014.

\bibitem{khorrami2015deep}
P.~Khorrami, T.~Paine, and T.~Huang, ``Do deep neural networks learn facial
  action units when doing expression recognition?'' in \emph{Proceedings of the
  IEEE International Conference on Computer Vision Workshops}, 2015, pp.
  19--27.

\bibitem{mehu2015emotion}
M.~Mehu and K.~R. Scherer, ``Emotion categories and dimensions in the facial
  communication of affect: An integrated approach.'' \emph{Emotion}, vol.~15,
  no.~6, p. 798, 2015.

\bibitem{liu2015faceattributes}
Z.~Liu, P.~Luo, X.~Wang, and X.~Tang, ``Deep learning face attributes in the
  wild,'' in \emph{Proceedings of International Conference on Computer Vision
  (ICCV)}, December 2015.

\bibitem{wang2019characterizing}
Z.~Wang, Z.~Dai, B.~P{\'o}czos, and J.~Carbonell, ``Characterizing and avoiding
  negative transfer,'' in \emph{Proceedings of the IEEE/CVF Conference on
  Computer Vision and Pattern Recognition}, 2019, pp. 11\,293--11\,302.

\bibitem{liu2019loss}
S.~Liu, Y.~Liang, and A.~Gitter, ``Loss-balanced task weighting to reduce
  negative transfer in multi-task learning,'' in \emph{Proceedings of the AAAI
  Conference on Artificial Intelligence}, vol.~33, no.~01, 2019, pp.
  9977--9978.

\bibitem{ruiz2015emotions}
A.~Ruiz, J.~Van~de Weijer, and X.~Binefa, ``From emotions to action units with
  hidden and semi-hidden-task learning,'' in \emph{Proceedings of the IEEE
  International Conference on Computer Vision}, 2015, pp. 3703--3711.

\bibitem{yang2016multiple}
J.~Yang, S.~Wu, S.~Wang, and Q.~Ji, ``Multiple facial action unit recognition
  enhanced by facial expressions,'' in \emph{2016 23rd International Conference
  on Pattern Recognition (ICPR)}.\hskip 1em plus 0.5em minus 0.4em\relax IEEE,
  2016, pp. 4089--4094.

\bibitem{wang2017expression}
S.~Wang, Q.~Gan, and Q.~Ji, ``Expression-assisted facial action unit
  recognition under incomplete au annotation,'' \emph{Pattern Recognition},
  vol.~61, pp. 78--91, 2017.

\bibitem{caruana1997multitask}
R.~Caruana, ``Multitask learning,'' \emph{Machine learning}, vol.~28, no.~1,
  pp. 41--75, 1997.

\bibitem{jang2019registration}
Y.~Jang, H.~Gunes, and I.~Patras, ``Registration-free face-ssd: Single shot
  analysis of smiles, facial attributes, and affect in the wild,''
  \emph{Computer Vision and Image Understanding}, vol. 182, pp. 17--29, 2019.

\bibitem{wang2017multi}
Z.~Wang, K.~He, Y.~Fu, R.~Feng, Y.-G. Jiang, and X.~Xue, ``Multi-task deep
  neural network for joint face recognition and facial attribute prediction,''
  in \emph{Proceedings of the 2017 ACM on International Conference on
  Multimedia Retrieval}.\hskip 1em plus 0.5em minus 0.4em\relax ACM, 2017, pp.
  365--374.

\bibitem{cui2020knowledge}
Z.~Cui, T.~Song, Y.~Wang, and Q.~Ji, ``Knowledge augmented deep neural networks
  for joint facial expression and action unit recognition,'' \emph{Advances in
  Neural Information Processing Systems}, vol.~33, 2020.

\bibitem{deng2020multitask}
D.~Deng, Z.~Chen, and B.~E. Shi, ``Multitask emotion recognition with
  incomplete labels,'' in \emph{2020 15th IEEE International Conference on
  Automatic Face and Gesture Recognition (FG 2020)(FG)}.\hskip 1em plus 0.5em
  minus 0.4em\relax IEEE Computer Society, 2020, pp. 828--835.

\bibitem{kollias2020analysing}
D.~Kollias, A.~Schulc, E.~Hajiyev, and S.~Zafeiriou, ``Analysing affective
  behavior in the first abaw 2020 competition,'' in \emph{2020 15th IEEE
  International Conference on Automatic Face and Gesture Recognition (FG
  2020)(FG)}.\hskip 1em plus 0.5em minus 0.4em\relax IEEE Computer Society, pp.
  794--800.

\bibitem{kuhnke2020two}
F.~Kuhnke, L.~Rumberg, and J.~Ostermann, ``Two-stream aural-visual affect
  analysis in the wild,'' in \emph{2020 15th IEEE International Conference on
  Automatic Face and Gesture Recognition (FG 2020)(FG)}.\hskip 1em plus 0.5em
  minus 0.4em\relax IEEE Computer Society, pp. 366--371.

\bibitem{wang2018two}
X.~Wang, M.~Peng, L.~Pan, M.~Hu, C.~Jin, and F.~Ren, ``Two-level attention with
  two-stage multi-task learning for facial emotion recognition,'' \emph{arXiv
  preprint arXiv:1811.12139}, 2018.

\bibitem{girard2017sayette}
J.~M. Girard, W.-S. Chu, L.~A. Jeni, and J.~F. Cohn, ``Sayette group formation
  task (gft) spontaneous facial expression database,'' in \emph{2017 12th IEEE
  International Conference on Automatic Face \& Gesture Recognition (FG
  2017)}.\hskip 1em plus 0.5em minus 0.4em\relax IEEE, 2017, pp. 581--588.

\bibitem{kossaifi2017afew}
J.~Kossaifi, G.~Tzimiropoulos, S.~Todorovic, and M.~Pantic, ``Afew-va database
  for valence and arousal estimation in-the-wild,'' \emph{Image and Vision
  Computing}, 2017.

\bibitem{li2017reliable}
S.~Li, W.~Deng, and J.~Du, ``Reliable crowdsourcing and deep
  locality-preserving learning for expression recognition in the wild,'' in
  \emph{Proceedings of the IEEE Conference on Computer Vision and Pattern
  Recognition}, 2017, pp. 2852--2861.

\bibitem{fabian2016emotionet}
C.~Fabian Benitez-Quiroz, R.~Srinivasan, and A.~M. Martinez, ``Emotionet: An
  accurate, real-time algorithm for the automatic annotation of a million
  facial expressions in the wild,'' in \emph{Proceedings of the IEEE Conference
  on Computer Vision and Pattern Recognition}, 2016, pp. 5562--5570.

\bibitem{benitez2017emotionet}
C.~F. Benitez-Quiroz, R.~Srinivasan, Q.~Feng, Y.~Wang, and A.~M. Martinez,
  ``Emotionet challenge: Recognition of facial expressions of emotion in the
  wild,'' \emph{arXiv preprint arXiv:1703.01210}, 2017.

\bibitem{mavadati2013disfa}
S.~M. Mavadati, M.~H. Mahoor, K.~Bartlett, P.~Trinh, and J.~F. Cohn, ``Disfa: A
  spontaneous facial action intensity database,'' \emph{Affective Computing,
  IEEE Transactions on}, vol.~4, no.~2, pp. 151--160, 2013.

\bibitem{zhang14bp4d}
X.~Zhang, L.~Yin, J.~F. Cohn, S.~Canavan, M.~Reale, A.~Horowitz, P.~Liu, and
  J.~M. Girard, ``Bp4d-spontaneous: a high-resolution spontaneous 3d dynamic
  facial expression database,'' \emph{Image and Vision Computing}, vol.~32,
  no.~10, pp. 692--706, 2014.

\bibitem{valstar2015fera}
M.~F. Valstar, T.~Almaev, J.~M. Girard, G.~McKeown, M.~Mehu, L.~Yin, M.~Pantic,
  and J.~F. Cohn, ``Fera 2015-second facial expression recognition and analysis
  challenge,'' in \emph{Automatic Face and Gesture Recognition (FG), 2015 11th
  IEEE International Conference and Workshops on}, vol.~6.\hskip 1em plus 0.5em
  minus 0.4em\relax IEEE, 2015, pp. 1--8.

\bibitem{zhang2016multimodal}
Z.~Zhang, J.~M. Girard, Y.~Wu, X.~Zhang, P.~Liu, U.~Ciftci, S.~Canavan,
  M.~Reale, A.~Horowitz, H.~Yang \emph{et~al.}, ``Multimodal spontaneous
  emotion corpus for human behavior analysis,'' in \emph{Proceedings of the
  IEEE Conference on Computer Vision and Pattern Recognition}, 2016, pp.
  3438--3446.

\bibitem{valstar2017fera}
M.~F. Valstar, E.~S{\'a}nchez-Lozano, J.~F. Cohn, L.~A. Jeni, J.~M. Girard,
  Z.~Zhang, L.~Yin, and M.~Pantic, ``Fera 2017-addressing head pose in the
  third facial expression recognition and analysis challenge,'' in \emph{2017
  12th IEEE International Conference on Automatic Face \& Gesture Recognition
  (FG 2017)}.\hskip 1em plus 0.5em minus 0.4em\relax IEEE, 2017, pp. 839--847.

\bibitem{najibi2017ssh}
M.~Najibi, P.~Samangouei, R.~Chellappa, and L.~Davis, ``{SSH}: Single stage
  headless face detector,'' in \emph{The IEEE International Conference on
  Computer Vision (ICCV)}, 2017.

\bibitem{yang2016wider}
S.~Yang, P.~Luo, C.~C. Loy, and X.~Tang, ``Wider face: A face detection
  benchmark,'' in \emph{IEEE Conference on Computer Vision and Pattern
  Recognition (CVPR)}, 2016.

\bibitem{parkhi2015deep}
O.~M. Parkhi, A.~Vedaldi, and A.~Zisserman, ``Deep face recognition.'' in
  \emph{British Machine Vision Conference ({BMVC})}, 2015.

\bibitem{he2016deep}
K.~He, X.~Zhang, S.~Ren, and J.~Sun, ``Deep residual learning for image
  recognition,'' in \emph{Proceedings of the IEEE Conference on Computer Vision
  and Pattern Recognition}, 2016, pp. 770--778.

\bibitem{huang2017densely}
G.~Huang, Z.~Liu, L.~Van Der~Maaten, and K.~Q. Weinberger, ``Densely connected
  convolutional networks,'' in \emph{Proceedings of the IEEE conference on
  computer vision and pattern recognition}, 2017, pp. 4700--4708.

\bibitem{kollias}
D.~Kollias, M.~A. Nicolaou, I.~Kotsia, G.~Zhao, and S.~Zafeiriou, ``Recognition
  of affect in the wild using deep neural networks,'' in \emph{Computer Vision
  and Pattern Recognition Workshops (CVPRW), 2017 IEEE Conference on}.\hskip
  1em plus 0.5em minus 0.4em\relax IEEE, 2017, pp. 1972--1979.

\bibitem{kollias2020va}
D.~Kollias and S.~Zafeiriou, ``Va-stargan: Continuous affect generation,'' in
  \emph{International Conference on Advanced Concepts for Intelligent Vision
  Systems}.\hskip 1em plus 0.5em minus 0.4em\relax Springer, 2020, pp.
  227--238.

\bibitem{wang2020region}
K.~Wang, X.~Peng, J.~Yang, D.~Meng, and Y.~Qiao, ``Region attention networks
  for pose and occlusion robust facial expression recognition,'' \emph{IEEE
  Transactions on Image Processing}, vol.~29, pp. 4057--4069, 2020.

\bibitem{ding2017facial}
W.~Ding, D.-Y. Huang, Z.~Chen, X.~Yu, and W.~Lin, ``Facial action recognition
  using very deep networks for highly imbalanced class distribution,'' in
  \emph{2017 Asia-Pacific Signal and Information Processing Association Annual
  Summit and Conference (APSIPA ASC)}.\hskip 1em plus 0.5em minus 0.4em\relax
  IEEE, 2017, pp. 1368--1372.

\bibitem{shao2018deep}
Z.~Shao, Z.~Liu, J.~Cai, and L.~Ma, ``Deep adaptive attention for joint facial
  action unit detection and face alignment,'' in \emph{Proceedings of the
  European Conference on Computer Vision (ECCV)}, 2018, pp. 705--720.

\bibitem{niu2019local}
X.~Niu, H.~Han, S.~Yang, Y.~Huang, and S.~Shan, ``Local relationship learning
  with person-specific shape regularization for facial action unit detection,''
  in \emph{Proceedings of the IEEE/CVF Conference on Computer Vision and
  Pattern Recognition}, 2019, pp. 11\,917--11\,926.

\bibitem{yuce2015discriminant}
A.~Y{\"u}ce, H.~Gao, and J.-P. Thiran, ``Discriminant multi-label manifold
  embedding for facial action unit detection,'' in \emph{2015 11th IEEE
  International Conference and Workshops on Automatic Face and Gesture
  Recognition (FG)}, vol.~6.\hskip 1em plus 0.5em minus 0.4em\relax IEEE, 2015,
  pp. 1--6.

\bibitem{tang2017view}
C.~Tang, W.~Zheng, J.~Yan, Q.~Li, Y.~Li, T.~Zhang, and Z.~Cui,
  ``View-independent facial action unit detection,'' in \emph{2017 12th IEEE
  International Conference on Automatic Face \& Gesture Recognition (FG
  2017)}.\hskip 1em plus 0.5em minus 0.4em\relax IEEE, 2017, pp. 878--882.

\bibitem{shao2019facial}
Z.~Shao, Z.~Liu, J.~Cai, Y.~Wu, and L.~Ma, ``Facial action unit detection using
  attention and relation learning,'' \emph{IEEE transactions on affective
  computing}, 2019.

\bibitem{yu2020gradient}
T.~Yu, S.~Kumar, A.~Gupta, S.~Levine, K.~Hausman, and C.~Finn, ``Gradient
  surgery for multi-task learning,'' \emph{Advances in Neural Information
  Processing Systems}, vol.~33, 2020.

\bibitem{lin2019pareto}
X.~Lin, H.-L. Zhen, Z.~Li, Q.~Zhang, and S.~Kwong, ``Pareto multi-task
  learning,'' in \emph{Thirty-third Conference on Neural Information Processing
  Systems (NeurIPS 2019)}, 2019.

\bibitem{wu2019understanding}
S.~Wu, H.~R. Zhang, and C.~R{\'e}, ``Understanding and improving information
  transfer in multi-task learning,'' in \emph{International Conference on
  Learning Representations}, 2019.

\bibitem{gendron2018universality}
M.~Gendron, C.~Crivelli, and L.~F. Barrett, ``Universality reconsidered:
  Diversity in making meaning of facial expressions,'' \emph{Current directions
  in psychological science}, vol.~27, no.~4, pp. 211--219, 2018.

\bibitem{bryant2019comparative}
D.~Bryant and A.~Howard, ``A comparative analysis of emotion-detecting ai
  systems with respect to algorithm performance and dataset diversity,'' in
  \emph{Proceedings of the 2019 AAAI/ACM Conference on AI, Ethics, and
  Society}, 2019, pp. 377--382.

\bibitem{han2017heterogeneous}
H.~Han, A.~K. Jain, F.~Wang, S.~Shan, and X.~Chen, ``Heterogeneous face
  attribute estimation: A deep multi-task learning approach,'' \emph{IEEE
  transactions on pattern analysis and machine intelligence}, vol.~40, no.~11,
  pp. 2597--2609, 2017.

\bibitem{hand2016attributes}
E.~M. Hand and R.~Chellappa, ``Attributes for improved attributes: A multi-task
  network for attribute classification,'' \emph{arXiv preprint
  arXiv:1604.07360}, 2016.

\end{thebibliography}

\begin{IEEEbiography}[{\includegraphics[width=1in,height=1.25in,clip,keepaspectratio]{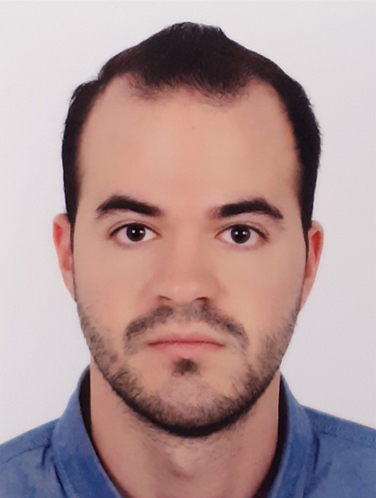}}]{Dimitrios Kollias}, Fellow of the Higher Education Academy, holder of a Post-Graduate Certificate and member of the IEEE, is currently a Senior Lecturer in Computer Science with the School of Computing and Mathematical Sciences, University of Greenwich. He has been the recipient of the prestigious Teaching Fellowship of Imperial College London. He has obtained the Ph.D. from the Department of Computing, Imperial College London, where he was a member of the iBUG group. Prior to this, he received the Diploma/M.Sc. in Electrical and Computer Engineering from the ECE School of the National Technical University of Athens, Greece, and the M.Sc. in Advanced Computing from the Department of Computing of Imperial College London. He has published his research in the top journals and conferences on machine learning, perception and computer vision such as IJCV, CVPR, ECCV, BMVC, IJCNN, ECAI and SSCI. He is a reviewer in top journals and conferences, such as CVPR, ECCV, ICCV, AAAI, TIP, TNNL, TAC, Neurocomputing, Pattern Recognition and Neural Networks.
He is the Chair in two the Competitions and Workshops in ICCV 2021. He  has been Competition Chair and Workshop Chair in IEEE FG 2020. He has won many grants and awards, such as from the City and Guilds College Association, the Imperial College Trust and the Complex \& Intelligent Systems Journal. He has h-index 18 and i10-index 19.  
His research interests span the areas of machine and deep learning, deep neural networks, computer vision, affective computing and medical imaging. 
\end{IEEEbiography}

\begin{IEEEbiography}[{\includegraphics[width=1in,height=1.25in,clip,keepaspectratio]{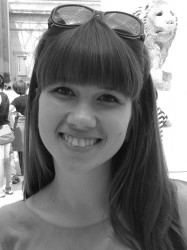}}]{Viktoriia Sharmanska}
Viktoriia Sharmanska is currently a Lecturer in AI at the Department of Informatics, University of Sussex, and an honorary lecturer at Imperial College London, UK. During 2017-2020, she was a recipient of a prestigious Imperial College Research Fellowship at the Department of Informatics, working on deep learning methods for human behavior analysis. 
Dr Sharmanska has co-authored numerous papers published at CVPR, ICCV/ECCV, NeurIPS, on novel statistical machine learning methodologies applied to computer vision problems, such as attribute-based object recognition, learning using privileged information, cross-modal learning, and recently on human facial behavior analysis and algorithmic fairness methods. 
She has built an international reputation such as being among the youngest Area Chair for top-tier international conferences in computer vision and deep learning such as ICLR since 2019, and CVPR 2021. Dr Sharmanska has received a number of prestigious awards, such as the Imperial College Research Fellowship 2017, Outstanding Reviewer Award at CVPR 2019. 
Her current research interests include deep learning methods for human behavior understanding from facial and bodily cues, video data synthesis, and algorithmic fairness methods to mitigate machine bias in visual data. 
\end{IEEEbiography}

\begin{IEEEbiography}[{\includegraphics[width=1in,height=1.25in,clip,keepaspectratio]{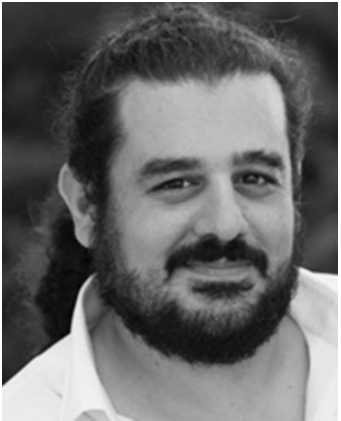}}]{Stefanos Zafeiriou} is currently a Professor in Machine Learning and Computer Vision with the Department of Computing, Imperial College London. He also holds an EPSRC Fellowship. He received the Prestigious Junior Research Fellowships from Imperial College London in 2011 to start his own independent research group. He received the President’s Medal for Excellence in Research Supervision for 2016. He received various awards during his doctoral and postdoctoral studies. He has been a Guest Editor of more than 6 journal special issues and co-organized more than 15 workshops/special sessions on specialized computer vision topics in top venues, such as CVPR/FG/ICCV/ECCV (including three very successfully challenges run in ICCV’13, ICCV’15 and CVPR'17 on facial landmark localisation/tracking). He has coauthored more than 70 journal papers mainly on novel statistical machine learning methodologies applied to computer vision problems, such as 2-D/3-D face analysis, deformable object fitting and
tracking, shape from shading, and human behavior analysis, published in the
most prestigious journals in his field of research, such as TPAMI,
IJCV, TIP, TNNLS and many papers in top conferences, such
as CVPR, ICCV, ECCV, ICML. His students are frequent recipients of very
prestigious and highly competitive fellowships, such as the Google, Intel and Qualcomm ones. He has more than
12000 citations to his work, h-index 54, i10-index	159. He was the General Chair of BMVC 2017.
\end{IEEEbiography}

\end{document}